\documentclass{article}

\PassOptionsToPackage{numbers, sort, compress}{natbib}

\usepackage[preprint]{neurips_2022}

\usepackage{subcaption,booktabs}
\usepackage{caption}

\usepackage[utf8]{inputenc} % allow utf-8 input
\usepackage[T1]{fontenc}    % use 8-bit T1 fonts
\usepackage{hyperref}       % hyperlinks
\usepackage{url}            % simple URL typesetting
\usepackage{booktabs}       % professional-quality tables
\usepackage{amsfonts}       % blackboard math symbols
\usepackage{nicefrac}       % compact symbols for 1/2, etc.
\usepackage{microtype}      % microtypography

\usepackage[table]{xcolor}
\usepackage{amsmath}
\usepackage{wrapfig}
\usepackage{multicol}        % used for the two-column index
\usepackage{multirow}
\usepackage{changepage}
\usepackage{makecell}
\usepackage{bbding}

\usepackage{amssymb}% http://ctan.org/pkg/amssymb
\usepackage{pifont}% http://ctan.org/pkg/pifont
\usepackage{array}
\newcolumntype{P}[1]{>{\centering\arraybackslash}p{#1}}
\usepackage{siunitx}
\usepackage{bbm}
\usepackage{dsfont}
\usepackage{caption} 
\usepackage{setspace}
\usepackage{balance}
\usepackage{bm} %数学公式加粗
\usepackage{color}
\usepackage{stfloats}
\usepackage{graphicx}
 \usepackage{mathrsfs}
\usepackage{booktabs} % For formal tables
\newcommand*\samethanks[1][\value{footnote}]{\footnotemark[#1]}
\usepackage{tikz}
\usetikzlibrary{decorations.pathreplacing,calc}
\newcommand{\tikzmark}[1]{\tikz[overlay,remember picture] \node (#1) {};}

\newcommand*{\AddNote}[4]{%
    \begin{tikzpicture}[overlay, remember picture]
        \draw [decoration={brace,amplitude=0.5em},decorate,ultra thick,black]
            ($(#3)!(#1.north)!($(#3)-(0,1)$)$) --  
            ($(#3)!(#2.south)!($(#3)-(0,1)$)$)
                node [align=center, text width=3cm, pos=0.5, anchor=west] {#4};
    \end{tikzpicture}
}% for big brace
\usepackage{floatrow}
\newfloatcommand{capbtabbox}{table}[][\FBwidth]

\usepackage[ruled,lined]{algorithm2e}
\usepackage{enumitem} 
\usepackage{ulem}

\title{NUWA-Infinity: Autoregressive over Autoregressive\\ Generation for Infinite Visual Synthesis}

\author{Chenfei Wu$^{1}$\thanks{Both authors contributed equally to this research.} \quad Jian Liang$^{2}$\samethanks[1] \quad Xiaowei Hu$^{3}$ \quad  Zhe Gan$^{3}$ \quad Jianfeng Wang$^{3}$\\ \textbf{Lijuan Wang}$^{3}$\quad  \textbf{Zicheng Liu}$^{3}$\quad  \textbf{Yuejian Fang}$^{2}$\quad \textbf{Nan Duan}$^{1}$\thanks{Corresponding author.} \\
 {$^{1}$Microsoft Research Asia \quad $^{2}$Peking University
 \quad $^{3}$Microsoft Azure AI} \\
{\tt\small\{chewu,xiaowei.hu,zhe.gan,jianfw,lijuanw,zliu,nanduan\}@microsoft.com}\\ {\tt\small\{j.liang@stu,fangyj@ss\}.pku.edu.cn}}

\begin{document}

\maketitle
\vspace{-9mm}

\begin{abstract}
\vspace{-4mm}
In this paper, we present NUWA-Infinity, a generative model for infinite visual synthesis, which is defined as the task of generating arbitrarily-sized high-resolution images or long-duration videos. An autoregressive over autoregressive generation mechanism is proposed to deal with this variable-size generation task, where a global patch-level autoregressive model considers the dependencies between patches, and a local token-level autoregressive model considers dependencies between visual tokens within each patch. A Nearby Context Pool (NCP) is introduced to cache-related patches already generated as the context for the current patch being generated, which can significantly save computation costs without sacrificing patch-level dependency modeling. An Arbitrary Direction Controller (ADC) is used to decide suitable generation orders for different visual synthesis tasks and learn order-aware positional embeddings. Compared to DALL·E, Imagen and Parti, NUWA-Infinity can generate high-resolution images with arbitrary sizes and support long-duration video generation additionally. Compared to NUWA, which also covers images and videos, NUWA-Infinity has superior visual synthesis capabilities in terms of resolution and variable-size generation. The GitHub link is \url{https://github.com/microsoft/NUWA}. The homepage link is \url{https://nuwa-infinity.microsoft.com}.

 %More creations from NUWA-Infinity can be found in \href{https://microsoft-research.nuwa}{https://microsoft-research.nuwa}.
 
 %Infinite visual synthesis aims to generate arbitrarily-sized images or videos. Most works tried to solve this task by first dividing images or videos into patches and then training the models on these patches without considering the dependencies between them. Thus, The quality and consistency are limited especially when generating large images or long videos. To address this issue, we propose NUWA-Infinity, an autoregressive over autoregressive pre-trained model for various visual synthesis tasks, with a global autoregression considering the dependency between patches and a local autoregression considering the dependency between visual tokens inside a patch. In detail, an Arbitrary Direction Modeling (ADM) considers different dependencies between patches by different orders, and enables the model to outpaint images into arbitrary wide and tall. A Nearby Context Pool (NCP) caches the context of patches during training, which enables the model to train one patch at each time and thus significantly saves computation without losing dependencies. Based on the above designs, NUWA-Infinity shows surprisingly good performance on extreme high-resolution images or long videos in five tasks.

\begin{figure}[htbp]
\vspace{-3mm}
    \centering
    \includegraphics[width=1\textwidth]{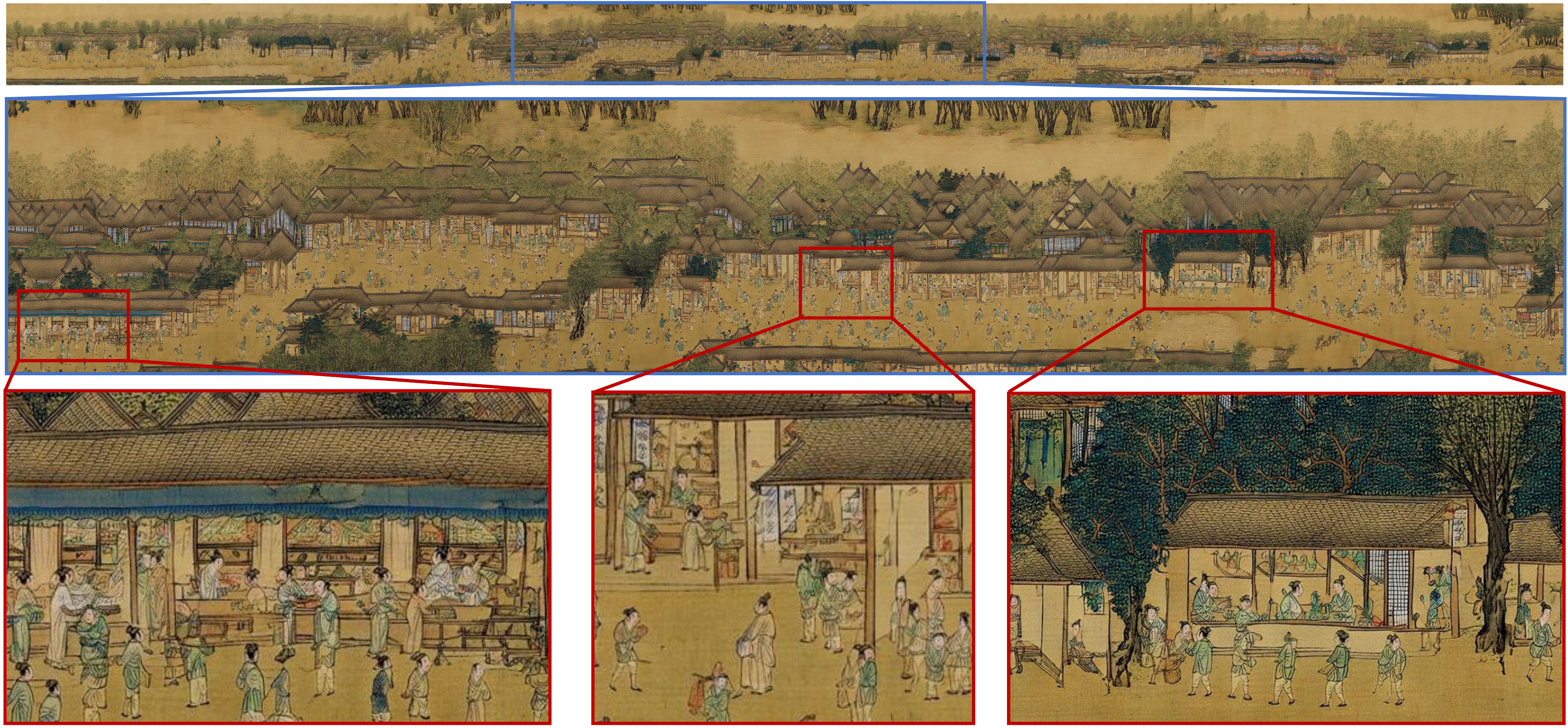}
    \caption{The painting at the top ($38912\times2048$) is created as Unconditional Image Generation\textsuperscript{\textbf{HD}} by NUWA-Infinity, which is trained on the famous painting \textit{Along the River During the Qingming Festival}. The patches in the middle or at the bottom highlight more details of this AI-created painting.}
    \label{fig:qm}
\end{figure}

% \begin{figure}[htbp]
%     \centering
%     \includegraphics[width=1\textwidth]{Figure/Tasks.pdf}
%     \caption{Four tasks supported by NUWA}
%     \label{fig:qm}
% \end{figure}

\begin{figure}[p]
 \vspace*{-2.3cm}
    \makebox[\linewidth]{
        \includegraphics[width=1.5\linewidth]{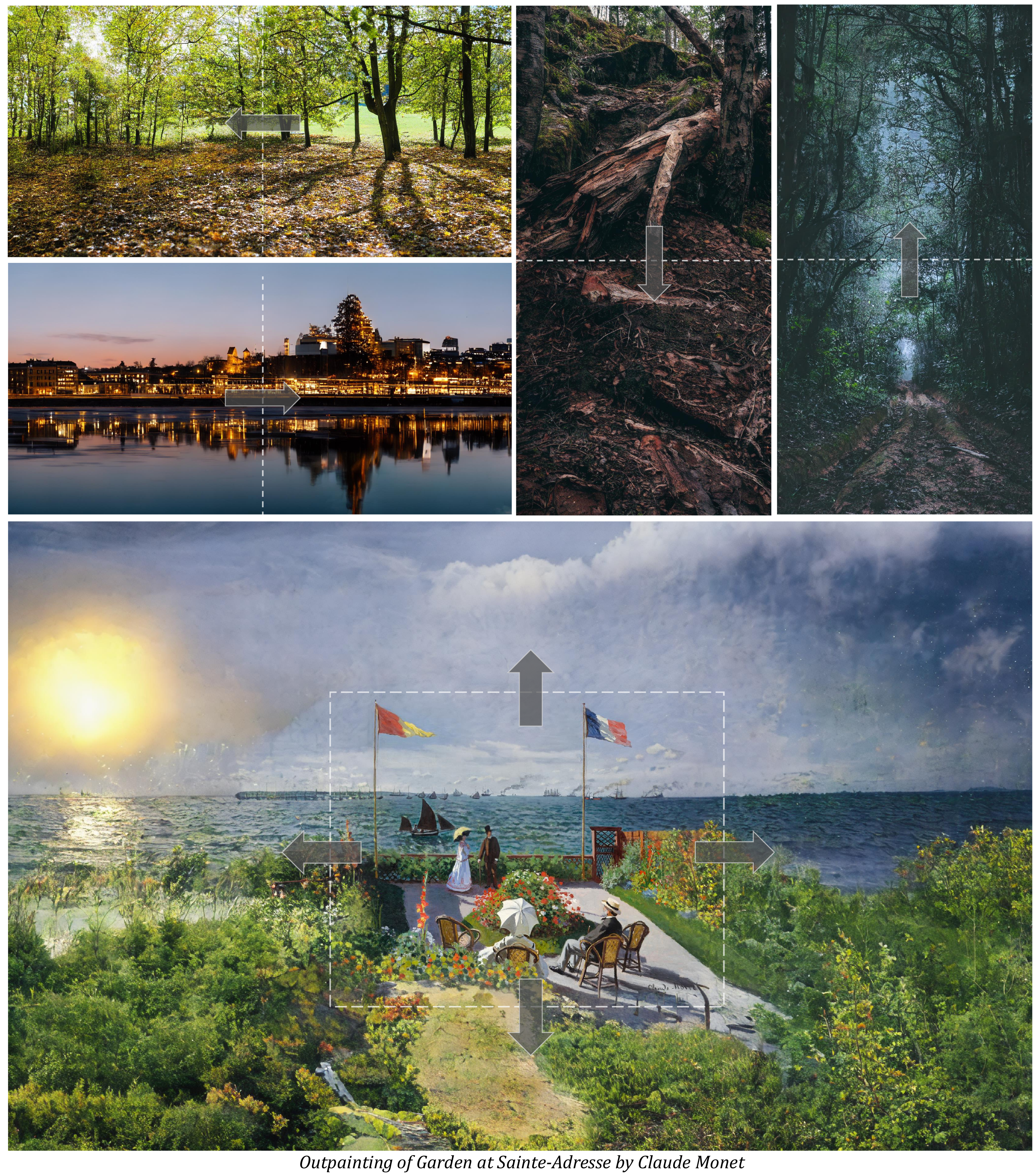}
    }
    \caption{Image Outpainting\textsuperscript{\textbf{HD}} examples. Four examples in the upper part ($2048 \times 1024$) show the outpainting of four different directions. The example in the lower part ($3328\times 2048$) shows outpainting in all directions of the famous painting of Monet, \textit{Garden at Sainte-Adresse}.}
\end{figure}

\begin{figure}[p]
 \vspace*{-2.3cm}
    \makebox[\linewidth]{
        \includegraphics[width=1.5\linewidth]{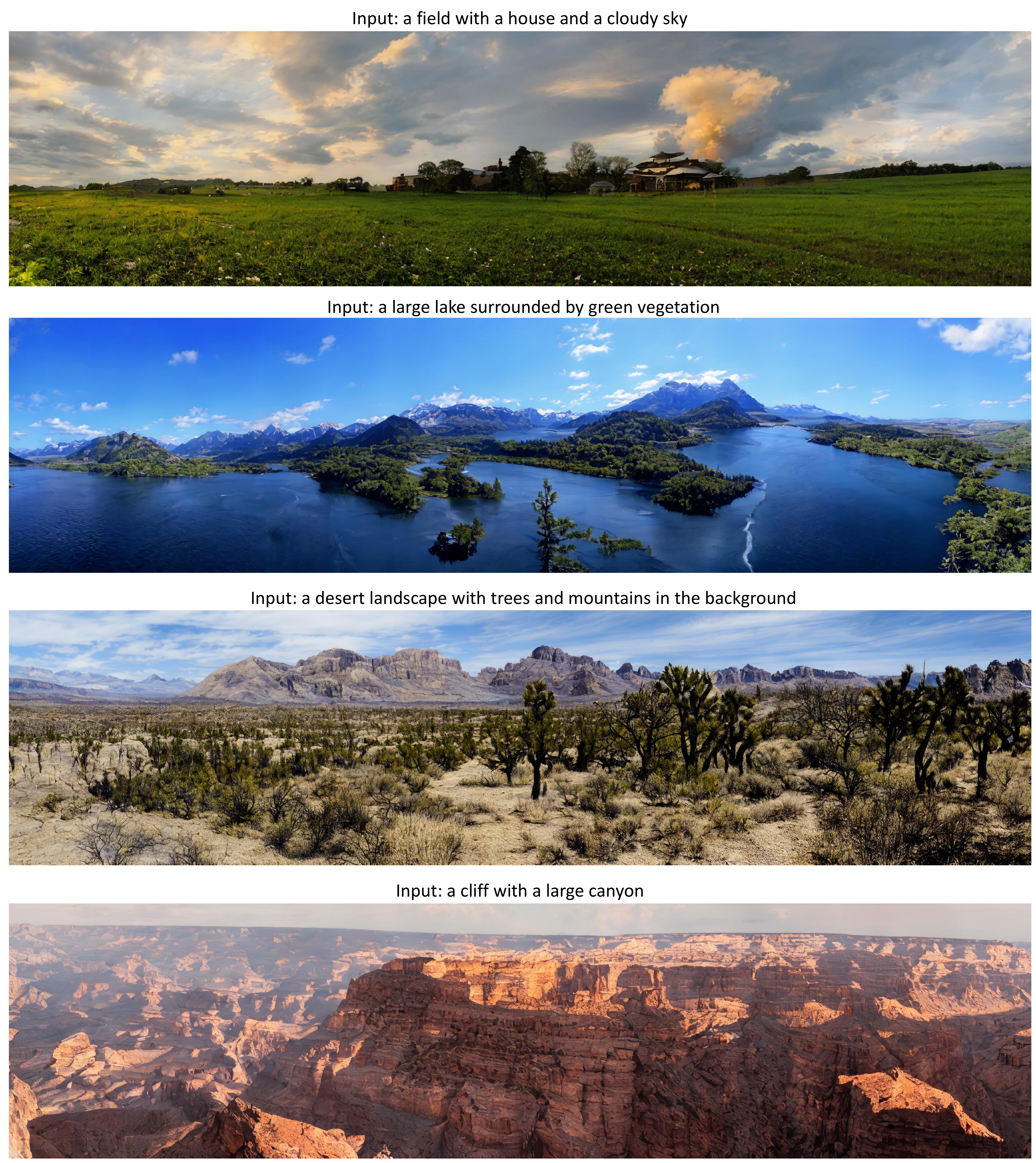}
    }
    \caption{Text-to-Image\textsuperscript{\textbf{HD}} examples in the resolution of $4096\times 1024$.  }
    \label{fig:tasks}
\end{figure}

\begin{figure}[p]
 \vspace*{-2.3cm}
    \makebox[\linewidth]{
        \includegraphics[width=1.5\linewidth]{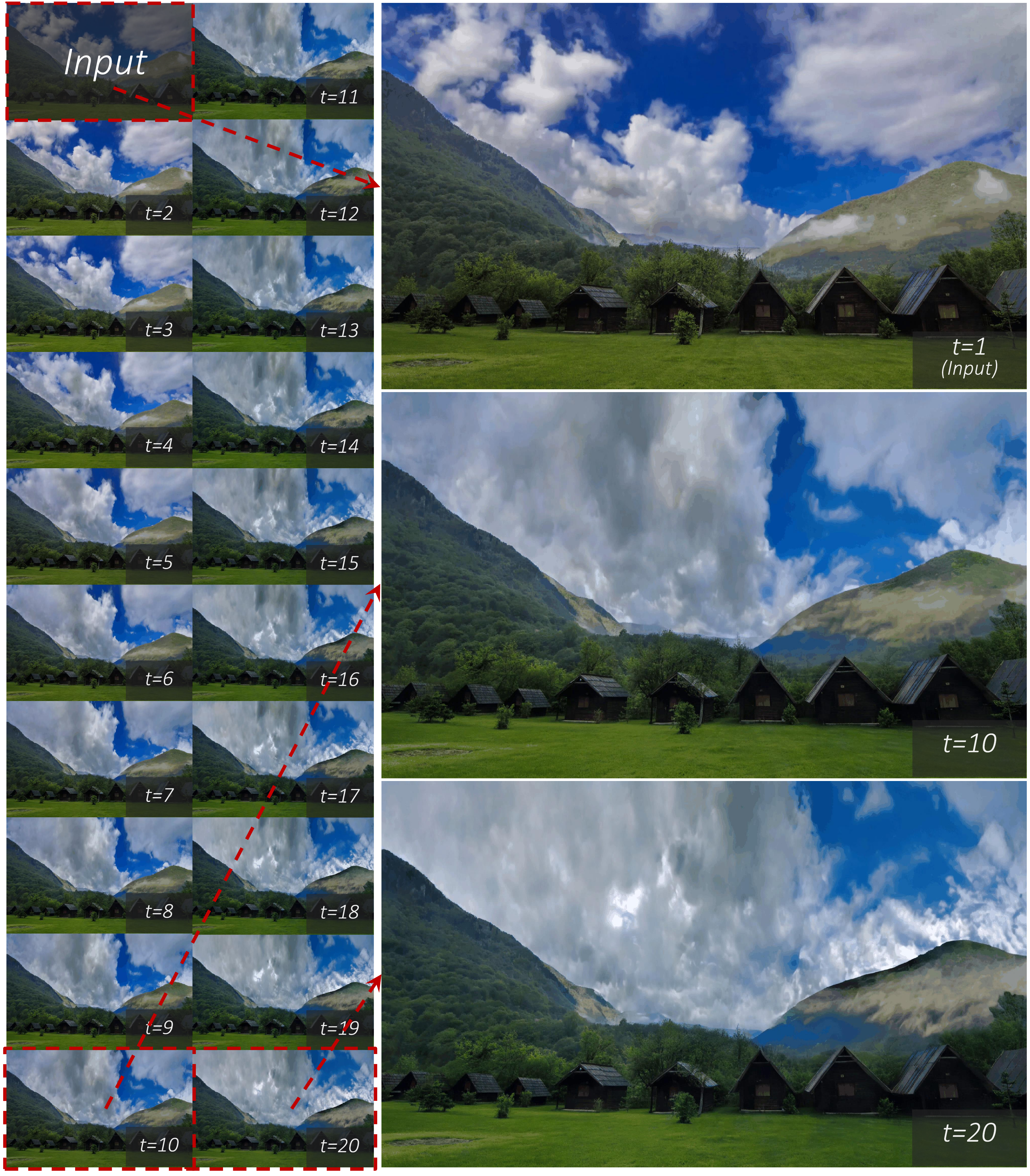}
    }
    \caption{Image Animation\textsuperscript{\textbf{HD}} examples in the resolution of $2560\times 1536$ with 20 frames.}
    \label{fig:ia_home}
\end{figure}

% \begin{figure}[p]
%  \vspace*{-2.3cm}
%     \makebox[\linewidth]{
%         \includegraphics[width=1.5\linewidth]{Figure/t2v_homeV2.pdf}
%     }
%     \caption{Text-to-Video\textsuperscript{\textbf{HD}} examples in the resolution of $1280 \times 1024$ with 8 frames. }
%     \label{fig:t2v_pig}
% \end{figure}

\end{abstract}

\section{Introduction}
\vspace{-2mm}
\newcommand{\chapquote}[3]{\begin{quotation} \textit{#1} \end{quotation} \begin{flushright} - #2, \textit{#3}\end{flushright} }
% \chapquote{``Nature is painting for us, day after day, pictures of infinite beauty.''}{John Ruskin}{(1819-1900)}

The ongoing convergence of vision-language representation and modeling techniques brings new opportunities to visual synthesis research. By learning visual knowledge and patterns from a large-scale visual and multimodal corpus, recent visual synthesis models can generate images or videos based on given text, visual, or multimodal inputs and support various visual content creation tasks, such as text-to-image or video generation, image inpainting or outpainting, video prediction, etc. We also witness a notable trend in this area, where more and more work begin to explore how to generate images with higher resolutions~\cite{dingCogViewMasteringTexttoImage2021,rameshHierarchicalTextConditionalImage2022,skorokhodovAligningLatentImage2021,linInfinityGANInfinitePixelImage2022,changMaskGITMaskedGenerative2022}, or generate videos with longer durations~\cite{villegasHighFidelityVideo2019,clarkAdversarialVideoGeneration2019,liuInfiniteNaturePerpetual2021}. This is because high-resolution images and long-duration videos can provide 
better visual effects to practical applications, such as design, advertisement, presentation, entertainment, etc.

However, generating arbitrarily-sized high-quality images (in resolution) or videos (in both resolution and duration) is not a trivial task. First, compared to text generation in the NLP field, the research of generating variable-size visual content is still in its early stage, and therefore is not well studied. Some existing works~\cite{struskiLocoGANLocallyConvolutional2022,skorokhodovAligningLatentImage2021,esserTamingTransformersHighResolution2021,changMaskGITMaskedGenerative2022} try to solve this problem with a divide-and-conquer strategy, which divides images or videos into patches, trains the model to generate patches separately without considering their dependencies, and composes the generated patches to form the final image or video. As such methods do not model the dependencies between the generated patches explicitly, they struggle to guarantee the consistency of generated contents, especially when generating high-resolution images or long-duration videos. 
Second, different from text generation that usually follows a fixed order (e.g., left-to-right), images have two dimensions (i.e., width and height), and videos have three (i.e., width, height, and duration). These suggest that visual synthesis models should consider and model different generation orders and directions for different types of tasks.

In this paper, we use \textbf{Infinite Visual Synthesis} to denote the task of generating arbitrarily-sized high-quality images or videos, and propose \textbf{NUWA-Infinity} as a general visual synthesis model that can solve the two challenges of this task mentioned before. First, NUWA-Infinity is based on an autoregressive over autoregressive generation mechanism, where a global patch-level autoregressive model considers the dependencies between patches, and a local token-level autoregressive model considers the dependencies between visual tokens within each patch. Compared to diffusion-based approaches~\cite{dhariwalDiffusionModelsBeat2021,hoDenoisingDiffusionProbabilistic2020,guVectorQuantizedDiffusion2022} that are only able to generate images with a fixed size, the autoregressive formulation naturally considers different levels of dependencies and perfectly deals with the variable-size generation task. We also introduce a Nearby Context Pool (NCP) to cache-related patches already generated as the context for the current patch being generated, which can significantly save the computation cost without sacrificing the patch-level dependency modeling. Second, we propose an Arbitrary Direction Controller (ADC) to decide suitable generation orders and learn order-aware position embeddings, which is extremely useful for image outpainting.

We evaluate NUWA-Infinity on five high-resolution visual synthesis tasks, including Unconditional Image Generation\textsuperscript{\textbf{HD}}, Text-to-Image\textsuperscript{\textbf{HD}}, Text-to-Video\textsuperscript{\textbf{HD}}, Image Animation\textsuperscript{\textbf{HD}} and Image Outpainting\textsuperscript{\textbf{HD}}. Compared to DALL·E~\cite{rameshZeroShotTexttoImageGeneration2021}, Imagen~\cite{sahariaPhotorealisticTexttoImageDiffusion2022} and Parti~\cite{yuScalingAutoregressiveModels2022}, which generate images with a fixed resolution (i.e., $1024\times1024$), NUWA-Infinity can generate high-resolution images with arbitrary sizes and support long-duration video generation additionally. Compared to NUWA~\cite{wuUWAVisualSynthesis2022}, which also supports image and video synthesis at the same time, the generation quality of NUWA-Infinity has been improved significantly. We also show the huge application potential of NUWA-Infinity on creative visual synthesis tasks, such as image outpainting and cartoon creation from natural language descriptions. We hope this technique can help visual content creators to save time, cut costs and improve their productivity and creativity.

\section{Related Work}\label{sec:relatedwork}
\vspace{-2mm}

%Both autoregressive and diffusion models have been deeply studied for visual content creation tasks.

\paragraph{Autoregressive Methods}
DALL·E \cite{rameshZeroShotTexttoImageGeneration2021} tokenizes each image into discrete visual tokens and trains an autoregressive model to generate visual tokens from the corresponding text. The output image is reconstructed by the VQVAE decoder, which takes the visual tokens generated by the autoregressive model as inputs.     
Parti \cite{yuScalingAutoregressiveModels2022} follows the same architecture of DALL·E, but uses ViT-VQGAN \cite{yuVectorquantizedImageModeling2022} to discretize and reconstruct images, which is an improved version of VQGAN from both architecture and codebook learning aspects.
NUWA \cite{wuUWAVisualSynthesis2022} is the first autoregressive visual synthesis pre-trained model to support both image and video generation tasks. 
Compared to these previous works, NUWA-Infinity introduces the autoregressive over autoregressive mechanism into the generation procedure, which enables the capability of generating variable-size images and videos. 

\paragraph{Diffusion Methods}
DALL·E 2 \cite{rameshHierarchicalTextConditionalImage2022} generates image embedding from an input text based on either an autoregressive or a diffusion model, and uses a diffusion model to produce the output image.
Imagen \cite{sahariaPhotorealisticTexttoImageDiffusion2022} uses a frozen large-scale pre-trained language model T5-XXL \cite{raffelExploringLimitsTransfer2020} to encode each input text, and uses two diffusion models to generate high-resolution images based on the text embeddings.
Both of these two diffusion-based text-to-image generation methods cannot support arbitrarily-sized image generation, as the size of the output images is pre-defined before training and inference.

%\paragraph{High-Resolution Visual Synthesis}
%High-resolution visual synthesis is a hot topic recently. Most models, such as Cogview~\cite{dingCogViewMasteringTexttoImage2021} and DALLE-2~\cite{ rameshHierarchicalTextConditionalImage2022}, firstly generate a small image, and then gradually scale it up. However, these two separate steps will result in a transmission gap between the small and the high-resolution images. Since a small image only contains a few limited objects, the super-resolution model can only scale it up, but cannot generate new ones. This paper focus on Infinite Visual Synthesis, which aims to generate images or videos of arbitrary size without super resolution.

\paragraph{Infinite Visual Synthesis}
To support infinite visual synthesis, most existing works follow the divide-and-conquer strategy to first divide a large image into several patches, and then train them in an independent way. GAN-based models~\cite{struskiLocoGANLocallyConvolutional2022,skorokhodovAligningLatentImage2021} attempt to divide large images into patches and optimize each of them from global or coordinated latent space independently. Since different patches have no explicit dependency, these models struggle to merge different patches during inference, and can easily lead to inconsistent results. To address this issue, autoregressive models~\cite{esserTamingTransformersHighResolution2021,changMaskGITMaskedGenerative2022} incorporate a sliding window to enforce dependencies between different patches during inference. Recently, Mask-Predict~\cite{choXLXMERTPaintCaption2020, zhangM6ufcUnifyingMultimodal2021, changMaskGITMaskedGenerative2022} also uses the sliding window approach, but incorporates a progressively mask-and-predict strategy to model dependencies between patches during the window sliding in inference. However, both autoregressive models and Mask-Predict models bring a huge gap between training and inference, since different patches are still trained independently but inferred in a dependent way. By introducing the autoregressive over autoregressive mechanism with Nearby Context Pool and Arbitrary Direction Controller, NUWA-Infinity can enable variable-size image and video generation, and save the computation cost without losing global dependency and consistency modeling.

\section{Model}\label{sec:method}
 \begin{figure}[tbp]
    \centering
    \includegraphics[width=\textwidth]{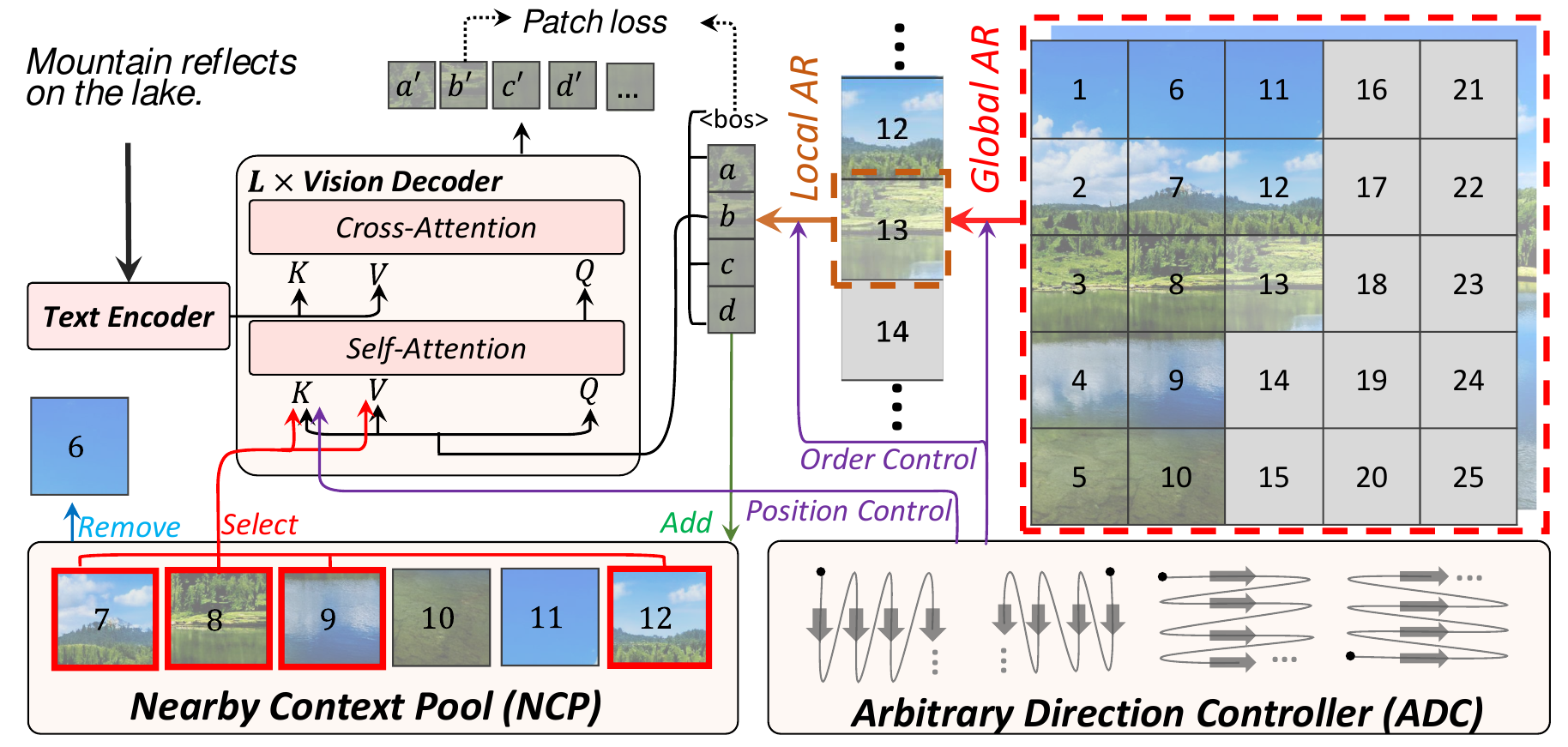}
    \caption{An overview of the proposed NUWA-Infinity model during the training process.}
    \label{fig:architecture}
\end{figure}

Given an input $y$, which can be a text or an image, the infinite visual synthesis task aims to generate an image or a video $x\in \mathbb{R}^{W\times H\times C \times F}$ with a user-specified resolution and duration, i.e., $\mathbb{P}(x|y)$, where $W$, $H$ and $C$ denote the width, height and channel of each image or video frame, respectively, $F$ denotes the frame number. $x$ denotes an image if $F=1$ and denotes a video if $F>1$. 

In general, NUWA-Infinity follows an autoregressive over autoregressive model to solve this task:
\begin{equation}\label{eq:pxy}
    \mathbb{P}\left(x|y \right )=\prod_{n=1}^{N}\mathbb{P}\left (p_{n}|p_{<n},y\right)
    =\prod_{n=1}^{N}\prod_{m=1}^{M}\mathbb{P}\left ( p_{n}^{(m)}|p_{<n},p_{n}^{(<m)}, y\right )
\end{equation}
$\prod_{n=1}^{N}\mathbb{P}\left (p_{n}|p_{<n},y\right)$ denotes the global autoregressive generation procedure, where $p_n$ is the $n^{th}$ patch being generated, $p_{<n}$ is the previous $n-1$ patches already generated, $N$ is the total number of patches. $\prod_{m=1}^{M}\mathbb{P}\left (p_{n}^{(m)}|p_{<n},p_{n}^{(<m)}, y\right )$ denotes the local autoregressive generation procedure, where $p_{n}^{(m)}$ is the $m^{th}$ visual token being generated in $p_n$, $p_{n}^{(<m)}$ is the previous $m-1$ visual tokens already generated, $M$ is the total number of visual tokens in each patch. Each $p_n$ is reconstructed by a pre-trained VQGAN decoder~\cite{esserTamingTransformersHighResolution2021}, which takes as input the visual token sequence $\{p_n^{(1)},...,p_n^{(M)}\}$. The final image or video is formed by composing all generated patches $\{p_1,...,p_N\}$ based on the specified resolution (i.e., $W \times H$) and duration (i.e., $F$).

NUWA-Infinity uses an encoder-decoder architecture to model the above generation procedure. In this paper, we mainly focus on five types of high-definition (\textbf{HD}) visual synthesis tasks, including Unconditional Image Generation\textsuperscript{\textbf{HD}}, Image Outpainting\textsuperscript{\textbf{HD}}, Image Animation\textsuperscript{\textbf{HD}}, Text-to-Image\textsuperscript{\textbf{HD}} and Text-to-Video\textsuperscript{\textbf{HD}}. In the $1^{st}$ task, the output image is generated by the vision decoder without any input. In the $2^{nd}$ and $3^{rd}$ tasks, the input image is fed into the vision decoder directly as the prefix to generate the output image or video. In the $4^{th}$ and $5^{th}$ tasks, the input text is encoded by the text encoder, and the output image or video is generated by the vision decoder.

One problem is that, different from text generation that follows the left-to-right order, images have two dimensions (i.e., width and height), and videos have three (i.e., width, height, and duration). These suggest the model should consider and handle different patch generation orders for different visual synthesis tasks. Motivated by this, we propose an Arbitrary Direction Controller (ADC) (Section~\ref{sec:adc}) that can plan proper patch generation orders and learn order-aware positional embeddings.

Another problem is that, the length (i.e., $N \times M$) of the visual tokens to be generated could be extremely long, which is challenging for most existing sequence generation models. To alleviate this issue, we propose a Nearby Context Pool (NCP) (Section~\ref{sec:ncp}) to cache-related patches already generated as the context for the current patch being generated, which can significantly save the computation cost without sacrificing the patch-level dependency modeling.

We train NUWA-Infinity (Section~\ref{sec:train}) using high-quality image-text pairs crawled from the web, and image-video pairs extracted from high-quality videos.

\subsection{Arbitrary Direction Controller (ADC)}\label{sec:adc}

In this subsection, we introduce Arbitrary Direction Controller (ADC), which provides two functions: $\textbf{Split}$, which splits images/videos and decides the patch generation order for training and inference procedures; $\textbf{Emb}$, which assigns order-aware positional embeddings based on the current context.

\begin{figure}[t!]
    \centering
    \includegraphics[width=1.0\textwidth]{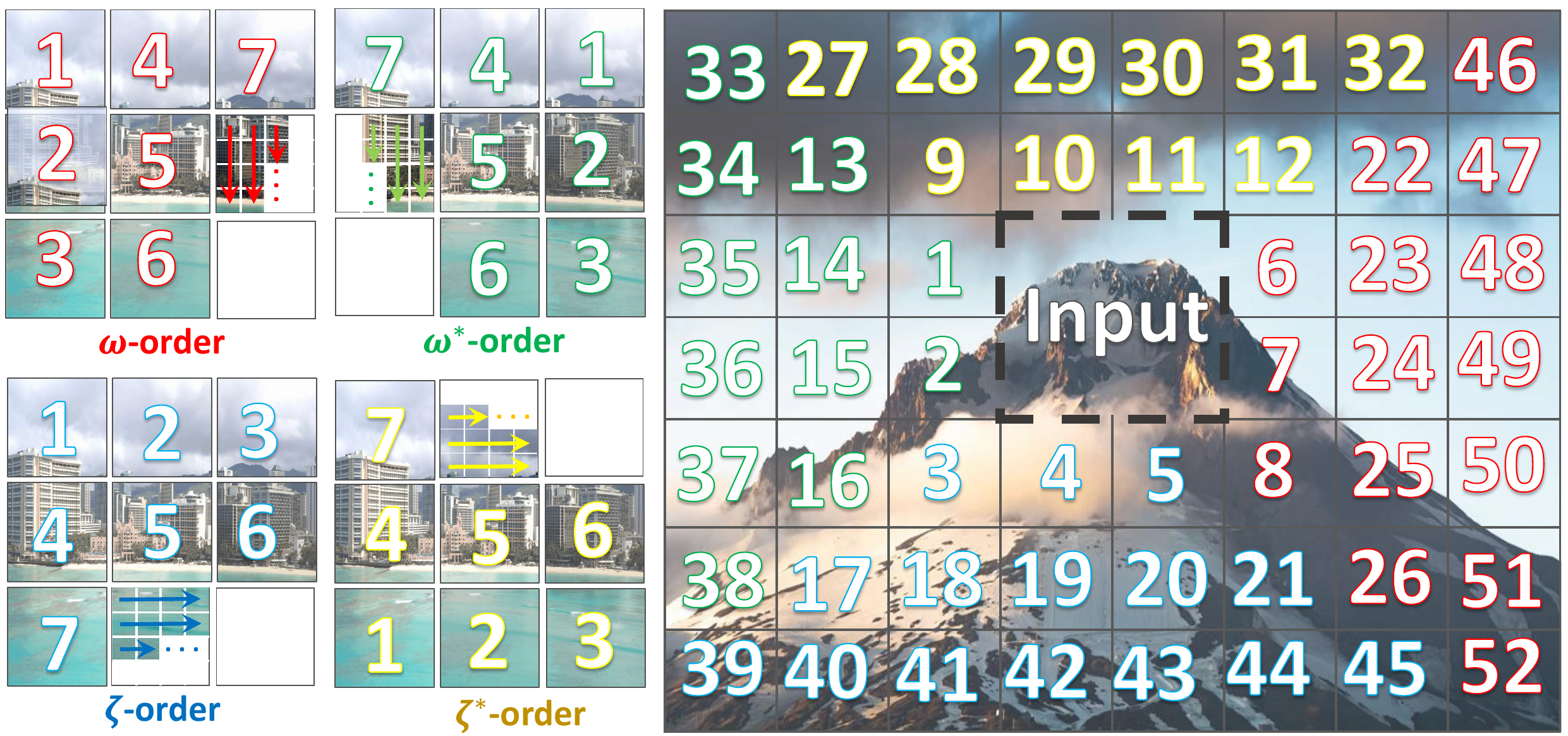}
    \caption{Illustration of patch order control in NUWA-Infinity. The left part shows four basic patch generation orders ($\omega, \omega^*,\zeta, \zeta^*$) during training. The right part shows how NUWA-Infinity performs the image outpainting task by composing these four orders. %\textcolor{red}{where the order obeys $\omega \rightarrow \zeta \rightarrow \omega^* \rightarrow \zeta^*$.}
    Arabic numerals indicate the order of global autoregression, arrows indicate the order of local autoregression.}
    \label{fig:adc}
    \vspace{-3mm}
\end{figure}
\begin{itemize}[leftmargin=*]
\item \textbf{Split}. This function takes the shape of an existing or to-be-generated image or video $x$ as input and returns an ordered patch sequence:
\begin{equation}\label{eq:split}
p_{1:N}=\text{ADC.Split}(x)
\end{equation}
$p_{1:N}=[p_1,...,p_N]$ denotes the ordered patch sequence. For simplicity, we use Fig.~\ref{fig:adc} to explain how this function splits an image into ordered patches in the training and inference stages. It is straightforward to extend from images to videos by considering the temporal dimension.

The left part of Fig.~\ref{fig:adc} shows how $\text{Split}(\cdot)$ works in the training stage. We define four basic generation orders and represent them using four Greek letters, respectively, according to their writing orders: $\omega$-order (↓→), $\omega^*$-order (↓←), $\zeta$-order (→↓), $\zeta^*$-order(→↑), where $*$ denotes the reversed writing order. When choosing the $\omega$-order in training, $\text{Split}(\cdot)$ will return patches in order of the red numerical sequence (top-left example in Fig.~\ref{fig:adc}). Similarly, the other three options will let NUWA-Infinity learn how to generate patches based on the corresponding orders. Note that there are more orders such as (↑←) or a snake-like order, but the above four basic orders and their compositions are enough to generate images in arbitrary resolutions or shapes.

The right part of Fig.~\ref{fig:adc} shows how $\text{Split}(\cdot)$ works in the inference stage for Image Outpainting\textsuperscript{\textbf{HD}}. Given a small image of a volcanic vent as input and its relative position in the targeted image with a specified larger resolution, the goal is to synthesize this targeted image by generating all the surrounding patches of the input. In order to leverage as many contextual patches as possible when generating new patches, 
$\text{Split}(\cdot)$ selects a patch generation order illustrated as a numerical sequence from $1$ to $52$. 

% Please note there could be multiple patch generation order options. 

\begin{figure}[t!]
    \centering
    \includegraphics[width=1.0\textwidth]{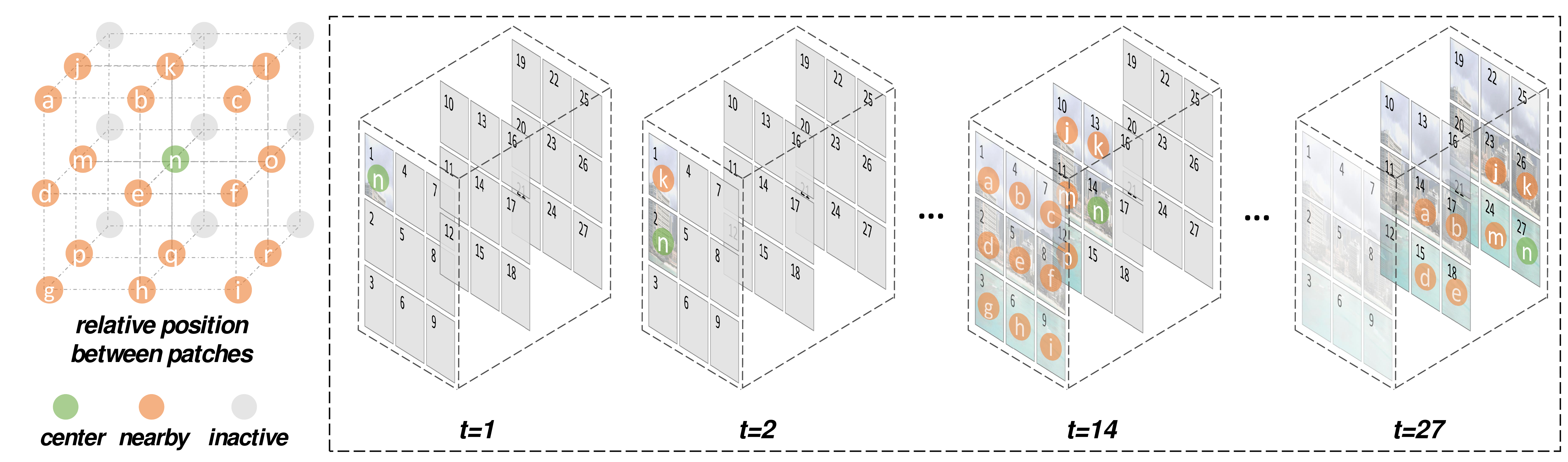}
    \caption{Illustration of dynamic position control in NUWA-Infinity. 
    %To make it simple, this example only uses 9 relative positions and generates 9 patches in $\omega$-order. For different patch generation orders, the patches will be dynamically assigned with different positional embeddings. 
    %If the patch generation order changes to $\omega^*, \zeta$ or $\zeta^*$, the position assignments will be different.
    }
    \label{fig:pos}
\end{figure}

\item \textbf{Emb}. This function assigns positional embeddings to the patch $p_n$ being generated and the patches in $c_n$ that are already generated and selected as the context of $p_n$:
\begin{equation}\label{eq:emb}
e_n=\text{ADC.Emb}([p_n; c_n])
\end{equation}

It is crucial to design a proper positional embedding for the order-aware patch generation procedure, since the absolute positional embedding \cite{vaswaniAttentionAllYou2017} is unable to consider and model all relative positions in image and video generation tasks. Motivated by the recent work on relative positional embedding~\cite{huangImproveTransformerModels2020,liuVideoSwinTransformer2022}, we propose a dynamic positional embedding in ADC, where dynamic means the positional embeddings could change according to different situations. Fig.~\ref{fig:pos} shows 18 dynamic relative embeddings from ``$a$'' to ``$r$''. The center patch being generated (in {\color{green}green} color) is always labeled as ``$n$'' and the relative positions of previously generated patches based on ``$n$'' are labeled by other symbols (in {\color{orange}orange} color). As a result, an embedding matrix of the size of $18\times d$ is formed. The right part shows how the embeddings are dynamically assigned to different patches when generating a specific patch. For example, when generating the last (i.e., $27^{th}$) patch, the positional embedding ``$n$'' is assigned to the current patch and the positional embeddings of ``$a$'',``$d$'',``$b$'',``$e$'',``$j$'',``$m$'',``$k$'' are assigned to patch 14, 15, 17, 18, 23, 24 and 26 in $c_{27}$, respectively.

% [question-2]: where is the comparison experiment that verifies the effectiveness of this technique?

% when training the first patch, the positional embedding of ``$\bullet$'' is assigned to patch 1 and thus ``$\bullet$'' is optimized. For the second patch, the positional embedding of ``$\bullet$'' is assigned to patch 2 and the positional embedding of ``↑'' is assigned to patch 1 and thus both of them are updated. Note that for step 3, we do not need to assign a positional embedding to patch 1, since the ``↑'' dependency between patch 1 and patch 2 have been used in step 2.

\end{itemize}

\subsection{Nearby Context Pool (NCP)}\label{sec:ncp}

\begin{figure}[htbp]
    \centering
    \includegraphics[width=1.0\textwidth]{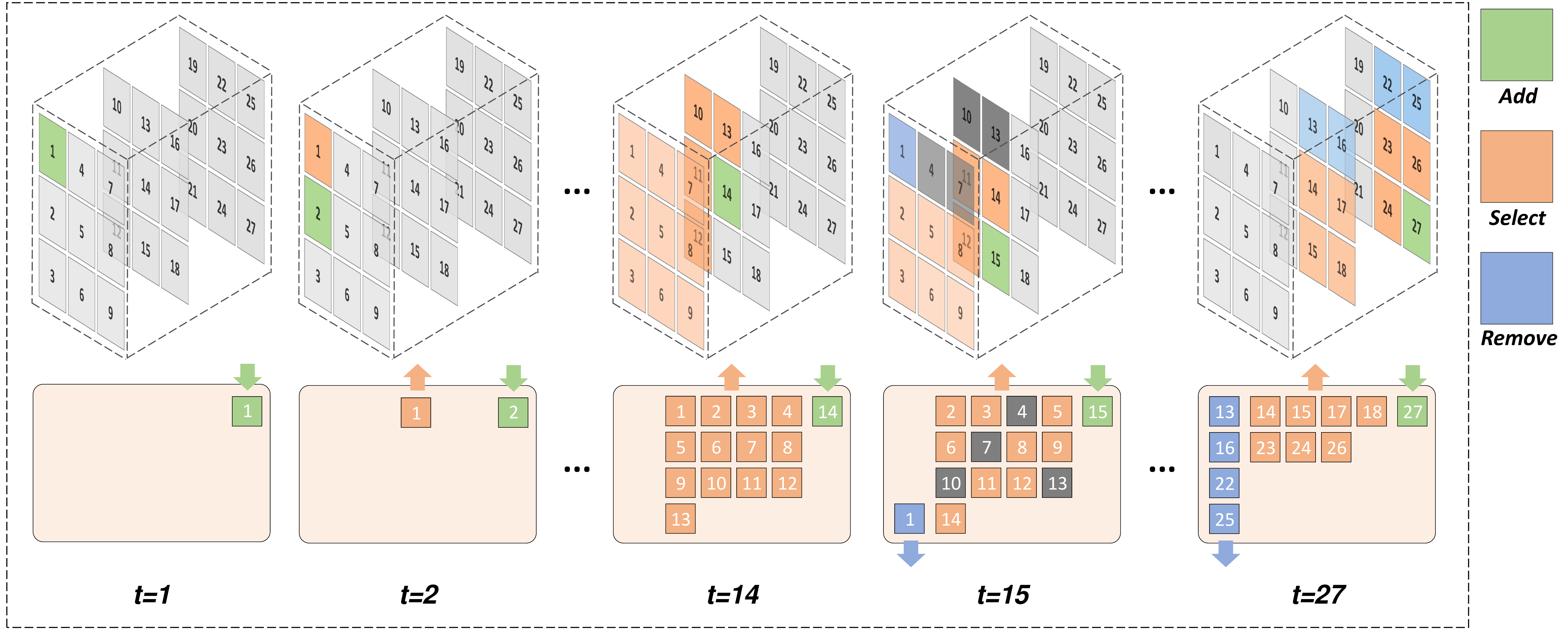}
    \caption{Illustration of NCP in $\omega$-order with a context extent of (1,1,1). 
    }
    % 27 frame.
    \label{fig:ncp_demo}
\end{figure}

An image or a video could be extremely large, thus the previous patches $p_{<n}$ in Eq.~\ref{eq:pxy} could have a large size. A natural idea is to only consider nearby patches as contexts. However, simply ignoring distant patches will lose long-term memory and thus harm the global consistency of the generated image or video. 
To address this issue, we propose a Nearby Context Pool (NCP) with three functions illustrated in Fig.~\ref{fig:ncp_demo}: 
\textbf{Select}, which dynamically selects nearby patches as the context to promote infinite generation; 
\textbf{Add}, which saves multi-layer hidden states of previously generated patches to help long-term memory; 
\textbf{Remove}, which removes expired caches for self-cleaning.

\begin{itemize}[leftmargin=*]

\item \textbf{Add}. This function adds the cache $a_n$ of the patch $p_n$ already generated into NCP:
\begin{equation}\label{eq:ncpadd}
\text{NCP.Add}(a_n)
\end{equation}
In NCP, the cache $a_n$ of the patch $p_n$ is defined as all the resulting multi-layer hidden states from the generation of $p_n$. Since NUWA-Infinity will not retain all generation history, this cache mechanism ensures the transmission of the necessary information in the whole generation procedure.
%As shown in Figure~\ref{fig:info}, since NUWA-Infinity is trained patch-by-patch, information of previous patches are transmitted into latter patches by the multi-layer hidden states stored in NCP. 
\item \textbf{Select}. This function selects the context $c_n$ for the patch $p_n$ to be generated:%.
\begin{equation}\label{eq:ncpselect}
c_n = \text{NCP.Select}(p_n)
\end{equation}
In NCP, the context $c_n$ of $p_n$ is defined as the caches of nearby patches already generated within a pre-defined 3D extent $(e^w, e^h, e^f)$, denoting the width extent, height extent, and frame extent. For example, when NUWA-Infinity generates the $14^{th}$ patch (as $p_n$) in Fig. \ref{fig:ncp_demo} and the maximum extent is set as $(1, 1, 1)$, the context will include the patches from 1 to 13.
%shows an extent of (1,1,1), which means the contexts contain one patch on the left, one patch on the right, one patch on the top, one patch on the bottom and one patch on the previous frame. We didn't consider next frames as context since temporal axis is uni-directional.

\item \textbf{Remove}. This function removes the caches of those patches in NCP that no longer have any effect on the generation of future patches:
\begin{equation}\label{eq:ncpremove}
\text{NCP.Remove}()
\end{equation}
In NCP, a cache is only cleaned when it cannot serve as the context for any patch to be generated. The $\text{Remove}(\cdot)$ function will be invoked after each $\text{Add}(\cdot)$ function.

\end{itemize}

\begin{wrapfigure}[12]{R}{0.5\textwidth}
\vspace{-0.17in}
\begin{minipage}{1\linewidth}
\begin{algorithm}[H]\label{alg:training}
\SetAlgoNoLine
\caption{Training Strategy}
\KwIn{images or videos $x$, optional text $y$}
\KwOut{optimized NUWA-Infinity model}

% Sample order $r$ from $\{\omega, \omega^*, \zeta, \zeta^*\}$ \;
Initial Arbitrary Direction Controller \textbf{ADC}\;
Initial Nearby Context Pool $ \textbf{NCP} \leftarrow \emptyset $ \;
%$x^{'}=\textbf{VQGANEncoder}(x)$\;
$p_{1:N} \leftarrow \textbf{ADC.Split}(x)$\;
\For{all $n$ from $1$ to $N$}{
    $c_n \leftarrow \textbf{NCP.Select}(p_n)$\;
    $e_n \leftarrow \textbf{ADC.Emb}([p_n; c_n])$\;
    $a_n,\hat{p}_n \leftarrow \textbf{NUWA}(p_n,c_n,e_n, y)$\;
    $\textbf{NCP.Add}(a_n)$\;
    $\textbf{NCP.Remove}()$\;
    $ \mathcal{L}_n=\text{CrossEntropy}(p_n,\hat{p}_n) $\;
    optimize $\mathcal{L}_n$\;
}
\end{algorithm}
\end{minipage}
\end{wrapfigure}

\subsection{Training and Inference Strategy}\label{sec:train}
This section will introduce the training and inference strategies of NUWA-Infinity in Algorithm~\ref{alg:training} and Algorithm~\ref{alg:inference}, respectively.

%%*************copy for fill the space
\subsubsection{Training Strategy}\label{sec:ts}

Given each input-output pair $\langle y,~x\in \mathbb{R}^{W\times H\times C\times F} \rangle$ in the pre-training corpus, we first split the visual data $x$ into patches and then randomly select one patch generation order $p_{1:N}=[p_1,...,p_N]$ from the four orders $\{\omega, \omega^*, \zeta, \zeta^*\}$ described in Section  \ref{sec:adc}. A pre-trained VQGAN encoder \cite{esserTamingTransformersHighResolution2021} transforms all images in $x$ into visual tokens $[p_1^{(1)},...,p_1^{(M)},...,p_N^{(1)},...,p_N^{(M)}]$, and each patch $p_n\in \mathbb{R}^{M\times d}$ is represented by its corresponding visual tokens $[p_n^{(1)},...,p_n^{(M)}]$. $y$ is encoded by a text encoder as $y'$, which denotes a sequence of token embeddings.

We train NUWA-Infinity based on each ordered patch sequence $r$. For the $n^{th}$ patch $p_n$, we first select its context $c_n\in \mathbb{R}^{N^c\times L \times M\times d}$ based on the NCP described in Section~\ref{sec:ncp}:
\begin{equation}\label{eq:x'}
c_n=\text{NCP}.\text{Select}(p_n)
\end{equation}
where $N^c$ denotes the number of patches in $q$, $L$ denotes the layer number of the vision decoder, $M$ denotes the visual token number in each context patch, and $d$ denotes the dimension of each visual token embedding. Note that $N^c$ can be changed during training, as different patches may have different numbers of context patches in $q$. The positional embeddings $e_n\in \mathbb{R}^{(1+N^c)\times d}$ of $p_n$ and its context $c_n$ are dynamically assigned by the ADC operation described in Section~\ref{sec:adc}:
\begin{equation}\label{eq:en}
e_n=\text{ADC.Emb}([p_n; c_n])
\end{equation}

Then, an $L$-layer vision decoder takes as input $p_n=[p_n^{(1)},...,p_n^{(M)}] \in \mathbb{R}^{M\times d}$ and $c_n$. In the $1^{st}$ layer, $p_n$ and the $1^{st}$ layer hidden states $c_n^{(1)}\in \mathbb{R}^{N^c\times M\times d}$ of all patches in $c_n$ are fed into a self-attention module, enhanced by the positional embeddings $e_n$:
\begin{equation}\label{eq:qkv}
\begin{split}
Q^s&=p_nW^q\\
K^s&=[p_n;c_n^{(1)}]W^k+e_n\\
V^s&=[p_n;c_n^{(1)}]W^v\\
\tilde{Q}^s&=\text{SelfAtt}(Q^s, K^s, V^s)  \\
\end{split}
\end{equation} 
$Q^s\in \mathbb{R}^{M\times d}, K^s\in \mathbb{R}^{(1+N^c)\times M\times d}, V^s\in \mathbb{R}^{(1+N^c)\times M\times d}$ are queries, keys and values, respectively, $W^q, W^k, W^v\in \mathbb{R}^{d\times d}$ are parameters to be learned, $\tilde{Q}^s$ denotes the attended results.

For the Text-to-Image\textsuperscript{\textbf{HD}} task, $\tilde{Q}^s$ and $y'$ are further fed into a cross-attention module as shown in Eq.~(\ref{eq:qkv2}).
\begin{equation}\label{eq:qkv2}
\begin{split}
Q^c&=\tilde{Q}^sW^{q'},\quad 
K^c=y'W^{k'}, \quad
V^c=y'W^{v'} \quad \\
\tilde{Q}^c&=\text{CrossAtt}(Q^c, K^c, V^c) \\
\end{split}
\end{equation}
where $\tilde{Q}^c\in \mathbb{R}^{M\times d}, K^c\in \mathbb{R}^{T\times d}, V^c\in \mathbb{R}^{T\times d}$ are queries, keys and values, respectively, $W^{q'}, W^{k'}, W^{v'} \in \mathbb{R}^{d\times d}$ are parameters to be learned, $T$ is the number of token embeddings in $y'$, $\tilde{Q}^c$ denotes the attended results.

By feeding $\tilde{Q}^s$ (for tasks without text input) or $\tilde{Q}^c$ (for tasks with text input) into a feed forward network, the output of the $1^{st}$ layer $\hat{p}_n^{(1)}\in \mathbb{R}^{M\times d}$ is obtained:
\begin{equation}\label{eq:pn1}
\hat{p}_n^{(1)}=\text{FFN}(\tilde{Q}^c)
\end{equation}

% Note that Eq.~(\ref{eq:qkv})$\sim$(\ref{eq:pn1}) only shows the $1^{st}$ layer of the visual decoder. In the following layer

% [p_n,\hat{p_n}^{(1)},\hat{p_n}^{(2)},.%..,\hat{p_n}^{(L-1)}])

By iteratively stacking Eq.~(\ref{eq:qkv})$\sim$(\ref{eq:pn1}) into $L$ layers, we obtain $\hat{p}_n^{(1)}, \hat{p}_n^{(2)}, ..., \hat{p}_n^{(L)}\in \mathbb{R}^{M\times d}$. $p_n$ and the previous $L-1$ layer outputs are concatenated to obtain a $L$-layer cache of the $n^{th}$ patch $p_n$:
\begin{equation}\label{eq:an}
a_n=[p_n;\hat{p}_n^{(1)};\hat{p}_n^{(2)};..;\hat{p}_n^{(L-1)}]
\end{equation}
where $a_n\in \mathbb{R}^{L\times M\times d}$. 
For simplicity, the procedure from Eq.~(\ref{eq:qkv}) to Eq.~(\ref{eq:an}) is defined as $\text{NUWA}$:
\begin{equation}\label{eq:nuwa}
\hat{p}_n, a_n=\text{NUWA}(p_n, c_n, e_n, y)
\end{equation}
where $\hat{p}_n=\hat{p}_n^{(L)}$ denotes the output embeddings. Then, NCP will collect the cache of $p_n$ to help the prediction of the next patches and conduct a self-cleaning to remove useless patches, as shown in Eq.~(\ref{eq:addremove}).
\begin{equation}\label{eq:addremove}
\begin{split}
    \text{NCP}.&\text{Add}(a_n)\\
    \text{NCP}.&\text{Remove}()\\
\end{split}
\end{equation}

Finally, the cross-entropy loss is used to optimize model parameters based on $\hat{p}_n$ and the ground-truth $p_n$.
%%%%%******************copy for full the space
% \begin{minipage}{0.4\textwidth}
% \begin{algorithm}[H]\label{alg:training}
% \SetAlgoNoLine
% \caption{Training Strategy}
% \KwIn{images or videos $x$, optional text $y$}
% \KwOut{optimized NUWA-Infinity model}

% % Sample order $r$ from $\{\omega, \omega^*, \zeta, \zeta^*\}$ \;
% Initial Arbitrary Direction Controller \textbf{ADC}\;
% Initial Nearby Context Pool $ \textbf{NCP} \leftarrow \emptyset $ \;
% %$x^{'}=\textbf{VQGANEncoder}(x)$\;
% $p_{1:N} \leftarrow \textbf{ADC.Split}(x)$\;
% \For{all $n$ from $1$ to $N$}{
%     $c_n \leftarrow \textbf{NCP.Select}(p_n)$\;
%     $e_n \leftarrow \textbf{ADC.Emb}([p_n; c_n])$\;
%     $a_n,\hat{p}_n \leftarrow \textbf{NUWA}(p_n,c_n,e_n, y)$\;
%     $\textbf{NCP.Add}(a_n)$\;
%     $\textbf{NCP.Remove}()$\;
%     $ \mathcal{L}_n=\text{CrossEntropy}(p_n,\hat{p}_n) $\;
%     optimize $\mathcal{L}_n$\;
% }
% \end{algorithm}
% \end{minipage}
%\hfill
% \vspace{-10mm}
\noindent\makebox[\textwidth][c]{
\begin{minipage}{0.75\textwidth}
\begin{algorithm}[H]\label{alg:inference}
\SetAlgoNoLine
\caption{Inference Strategy}

\KwIn{a text $y$ or an image $h$}
\KwOut{generated image/video $x$}

Initial Arbitrary Direction Controller \textbf{ADC}\; 
Initial Nearby Context Pool $ \textbf{NCP} \leftarrow$ \tikzmark{right1} $ \emptyset $ \ ;
$q_{1:K} \leftarrow \tikzmark{right2} \textbf{ADC.Split}(h)$\; 
\For{all $k$ from $1$ to $K$ \tikzmark{top1}}
{
    $c_k \leftarrow \textbf{NCP.Select}(q_k)$\;
    $e_k \leftarrow \textbf{ADC.Emb}([q_k; c_k])$\;
    $a_k,\hat{q}_k \leftarrow \textbf{NUWA}(q_k, c_k,e_k, y)$\;
    $\textbf{NCP.Add}(a_k)$\;
    $\textbf{NCP.Remove}()$\;
    \tikzmark{bottom1}
}

%Initial $ x' \leftarrow [0]_s $\;
%$h'=\textbf{VQGANEncoder}(h)$\;
%$p_{1:N},q_{1:K} \leftarrow \textbf{ADC.Split}(x', h')$\;
%$\bm{e} \leftarrow \textbf{ADC.Emb}(q_{1:K})$\;

%\For{all $k$ from 1 to K}{
%$a_k,\hat{q}_k \leftarrow %\textbf{NUWA}(q_k,\emptyset,\bm{e},y)$%\;
%}
%Initial Nearby Context Pool $ \textbf{NCP} \leftarrow \bm{a} $ \;

% Initial $n \leftarrow 1 $\;
$p_{1:N} \leftarrow \textbf{ADC.Split}(x)$\; 
\For{all $n$ from 1 to $N$}{
    $c_n \leftarrow \textbf{NCP.Select}(p_n)$\;
    $e_n \leftarrow \textbf{ADC.Emb}([p_n; c_n])$\;
    $a_n,\hat{p}_n \leftarrow \textbf{NUWA}(\emptyset, c_n,e_n,y)$\;
    $\textbf{NCP.Add}(a_n)$\;
    $\textbf{NCP.Remove}()$\;
    $x' \leftarrow [x';\hat{p}_n]$\;
}
\eIf{target $x$ is an image \tikzmark{top2}}
{return $\textbf{VQGANDecoder}(x')$}
{return $\textbf{PixelGuidedVQGANDecoder}(x')$ \tikzmark{bottom2}}

\AddNote{top1}{bottom1}{right1}{Image Condition Pre-caching}
\AddNote{top2}{bottom2}{right2}{Pixel-Guided VQGAN}
\end{algorithm}
\vspace{-3mm}
\end{minipage}
}

\subsubsection{Inference Strategy}\label{sec:is}
\vspace{-3mm}
NUWA-Infinity can support various visual synthesis scenarios, and we focus on five tasks in this paper: Unconditional Image Generation\textsuperscript{\textbf{HD}}, Image Outpainting\textsuperscript{\textbf{HD}}, Image Animation\textsuperscript{\textbf{HD}}, Text-to-Image\textsuperscript{\textbf{HD}} and Text-to-Video\textsuperscript{\textbf{HD}}. There are two specific designs for the last four tasks: Image Condition Pre-caching and Pixel-Guided VQGAN.

\paragraph{Image Condition Pre-caching}
\vspace{-3mm}
For Image Outpainting\textsuperscript{\textbf{HD}} and Image Animation\textsuperscript{\textbf{HD}}, the input is an image condition $h$
%and an optional text 
and the output is a spatial extended image or a temporal extended video. 
%To handle the input image condition $a$, 
%We propose an Image Condition Pre-caching mechanism for these two tasks. Firstly, 
A VQGAN encoder is used to encode $h$ into a list of patches with corresponding visual tokens, where $K$ denotes the number of conditional patches. Then, these patches and visual tokens are fed into the vision decoder (Eq.~\ref{eq:en}$\sim$\ref{eq:addremove}) as a prefix to initialize NCP, which will be used next to generate the extended image or following video frames.

\paragraph{Pixel-Guided VQGAN}
\vspace{-3mm}
\begin{wrapfigure}{r}{0.4\textwidth}
\vspace{-7mm}%-7
    \includegraphics[width=0.8\textwidth]{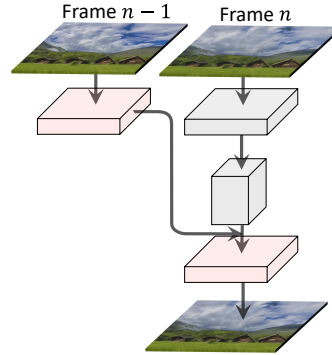}
    \caption{Pixel-Guided VQGAN.}
    \label{fig:pgvqgan}
\end{wrapfigure}

For Image Animation\textsuperscript{\textbf{HD}} and Text-to-Video\textsuperscript{\textbf{HD}}, the output are videos. Since traditional VQVAE is trained only on images, simply decoding a video frame-by-frame will lead to inconsistency between frames. To solve this issue, we propose a Pixel-Guided VQGAN (PG-VQGAN), as shown in Fig.~\ref{fig:pgvqgan}. Different from traditional VQGAN which is trained on images independently, we sample 2 consecutive frames $n-1$ and $n$ as a training pair and use the pixel-level information of the $n-1$ frame to enhance the decoder of frame $n$. In detail, the frame $n-1$ is encoded with the same number of layers as the traditional VQGAN decoder, and the output of each encoder layer is fused with the corresponding output layer of the VQGAN decoder. We simply use an element-sum operation as the fusion strategy and observe promising results (see Fig.~\ref{fig:pgvqgan}). 
When decoding the first frame during inference, Image Animation\textsuperscript{\textbf{HD}} has a ground-truth $n-1$ frame. However, 
For Text-to-Video\textsuperscript{\textbf{HD}}, since there are no ground-truth frames, we instead use traditional VQGAN to decode the first frame and Pixel-Guided VQGAN to decode the following frames.

%  \begin{figure}[tbp]
%     \centering
%     \includegraphics[width=0.5\textwidth]{Figure/PGVQGAN.pdf}
%     \caption{Inference pipeline for downstream tasks for NUWA-Infinity.}
%     \label{fig:inference}
% \end{figure}

 \begin{figure}[tbp]
    \centering
    \includegraphics[width=\textwidth]{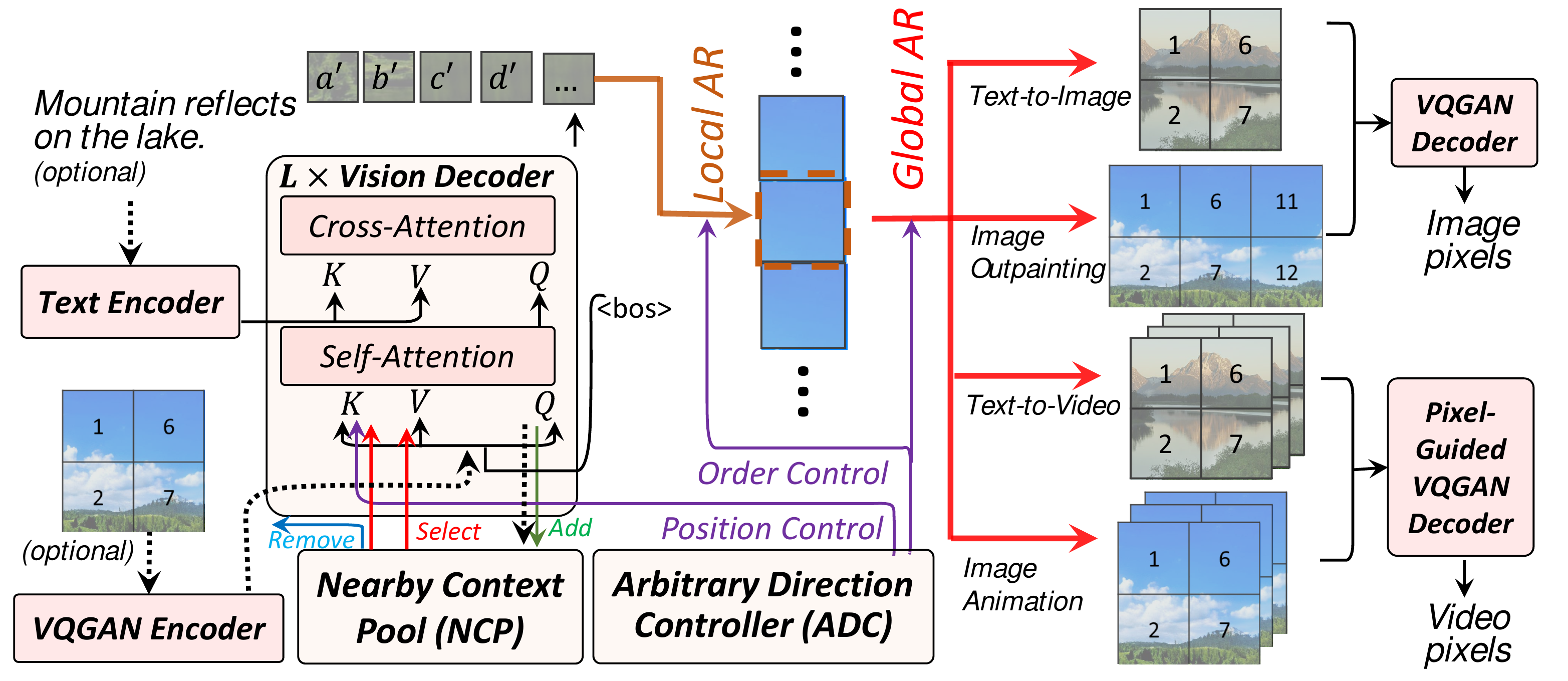}
    \caption{Inference pipeline of NUWA-Infinity for downstream tasks.}
    \label{fig:inference}
    \vspace{-4mm}
\end{figure}

\section{Experiments}
\vspace{-2mm}
\subsection{Datasets}
\vspace{-2mm}
Different from most visual synthesis works, NUWA-Infinity focuses on generating images and videos with high resolutions and long durations. As a result, most existing datasets cannot be used in training or evaluation. To evaluate the ability of NUWA-Infinity on the five tasks mentioned before, we build four datasets with high resolutions ($\geqslant 1024^{2}$) in the following:
\begin{itemize}[leftmargin=*]

\item \textbf{RQF}. We build a new dataset, Riverside of Qingming Festival (RQF), based on the version of \emph{\textit{Along the River During the Qingming Festival}} drawn by Qiu Ying\footnote{https://www.comuseum.com/painting/masters/qiu-ying/}. This version is 30.5 centimeters high and 987 centimeters wide in reality, and we download a digital print with $4200\times 135912$ pixels. We resize the whole image to $2048\times 66270$ resolution and split it into several overlapped $2048\times 2048$ patches instead of non-overlapping ones. We end with a dataset of 128 images.  We train NUWA-Infinity with the $<\emptyset, image>$ pairs on the dataset with $2048\times 2048$ resolution but qualitatively evaluate Unconditional Image Generation\textsuperscript{\textbf{HD}} at higher resolution $2048\times 38912$.

\item \textbf{LHQC}. We build a new dataset, Landscape High Quality with Captions (LHQC), based on the publicly available dataset LHQ~\cite{skorokhodovAligningLatentImage2021}. The original LHQ dataset consists of $90$K high-resolution ($\geqslant 1024^{2}$) nature landscapes. To support the text prompt, we use the image captioning model from~\cite{wangGITGenerativeImagetotext2022} to generate the captions for the dataset first and manually fix some errors in the generated results. We finally obtain a dataset with $90$K text-image pairs. We spilt the dataset into two splits: train ($85$K) and test ($5$K). We train NUWA-Infinity with the $<text, image>$ pairs on the train split and evaluate Text-to-Image\textsuperscript{\textbf{HD}} and Image Outpainting\textsuperscript{\textbf{HD}} on the test split.

\item \textbf{LHQ-V}. We build a new dataset, Landscape High Quality for Videos (LHQ-V), based on the videos scrapped from the www.pexels.com website. We first build a query set with 100 landscape related keywords (e.g., ``sky'', ``forest'', ``cloud''). Then, we query the website using these keywords and obtain $85$K high-resolution videos. To further make the dataset cleaner, we use Mask R-CNN \cite{heMaskRCNN2017} to detect objects from these videos and remove videos containing objects that are not related to landscape (i.e., ``table'', ``human'', ``computer''). Finally, we obtain a dataset with $40$K videos. We split the dataset into two splits: train ($38$K) and test ($2$K). We train NUWA-Infinity with the $<\emptyset, video>$ pairs on the train split and evaluate Image Animation\textsuperscript{\textbf{HD}} on the test split.

\item \textbf{PeppaPig}. We build a new dataset, PeppaPig, based on the famous cartoon PeppaPig. We collect Season 1-4 videos of PeppaPig and split the videos into multiple clips based on the timeline of subtitles. We then ask 20 trained annotators to annotate the captions for these videos. If the clip is not smooth, the annotators are asked to provide a ``N/A'' caption. We also ask a meta annotator to check these captions. Finally, we remove videos with the ``N/A'' caption and obtain $10$K text-video pairs. We split the dataset into two splits: train ($9$K) and test ($1$K). We train NUWA-Infinity with the $<text, video>$ pairs on the train split and evaluate Text-to-Video\textsuperscript{\textbf{HD}} on the test split.
\end{itemize}

\begin{table}[t]
\small
\centering
\setlength{\tabcolsep}{5pt}
\begin{tabular}{lcccc}
\toprule
Setting                      & UIG\textsuperscript{\textbf{HD}} & T2I\textsuperscript{\textbf{HD}} \& IO\textsuperscript{\textbf{HD}} & IA\textsuperscript{\textbf{HD}} & T2V\textsuperscript{\textbf{HD}} \\
\midrule
\multicolumn{5}{l}{\textit{VQVAE}}
\\ Backbone & VQGAN & VQGAN & PG-VQGAN & PG-VQGAN
\\ 
Codebook               & 16384      &16384         & 16384     & 16384           \\
Dimension              & 256         & 256         & 256        & 256           \\
Compression Ratio      & 16          & 16        & 16           & 16        \\
\midrule
\multicolumn{5}{l}{\textit{Transformer}}\\
Layer Number $L$    & 24                  & 24                   & 24                  & 24\\
Hidden Dimension $d$ & 1280                & 1280                 & 1280                & 1280                 \\
Head Number     & 20                  & 20                   & 20                  & 20                   \\
Self-attention   & \checkmark                 & \checkmark                   & \checkmark                 & \checkmark                   \\
Cross-attention  & \XSolidBrush                & \checkmark               & \XSolidBrush                    & \checkmark               \\
%\midrule
%\multicolumn{5}{l}{\textit{Patch}}\\
%Size                   & 256                 & 256                  & 256                 & 256                  \\
\midrule
\multicolumn{5}{l}{\textit{NCP \& ADC}}\\
Patch Number $N$ & 64 & 16 & 80 & 80\\
Patch Tokens $M$ & 256 & 256 & 256 & 256\\
%Sequence Length $N^c$ & ? & ? & ? & ?\\
Context Extent       & (2, 2, 0)       & (2, 2, 0)               & (1, 1, 3)       & (2, 2, 3)              \\
\midrule
\multicolumn{5}{l}{\textit{Dataset}}\\
Name                      & RQF                & LHQC               & LHQ-V                & PeppaPig                \\
Train Scale              & 128                 & 85K                  & 38K                 & 10K                  \\
Test Scale                  & N/A                  & 5K                    & 2K                  & 1K                   \\
\midrule
\multicolumn{5}{l}{\textit{Training \& Inference}}\\
Epoch               & 6000                  & 50                   & 50                  & 150                   \\
Visual Size $W \times H$            & $2048\times 2048$         & $1024\times 1024$          & $1024\times 1024$         & $1024\times 1024$          \\
Frame Length $F$ & N/A & N/A & 5 & 5 \\
Text Length              & N/A              & 77                 & N/A               & 77                 \\
Batch Size                   & 128                 & 512                  & 256                 & 256                 \\
Learning Rate                &$1\times 10^{-4}$                & $1\times 10^{-4}$                 &$1\times 10^{-4}$                & $1\times 10^{-4}$                 \\
Warmup Ratio                 & 2\%                 & 5\%                 & 5\%                 & 5\%                 \\
\bottomrule
\end{tabular}
\caption{Implementation details of model training for different tasks. %for large model, the base model has fewer parameters than the large model, it uses 16 layers, 768 hidden hidden dimensions and 12 heads. 
%We use Pixel-guided VQGAN (PG-VQGAN) instead of common VQGAN for Image Animation\textsuperscript{\textbf{HD}} and Text-to-Video\textsuperscript{\textbf{HD}}.
}
\label{tab:implementation_details}
\end{table}

\subsection{Implementation Details}\label{sub:imp}

The training of NUWA-Infinity can be split into two stages. In the first stage, all images are cropped into $1024 \times 1024$ and videos are cut into $1024 \times 1024 \times 5$ with $5$ fps. Then, they are encoded into discrete visual tokens using the VQGAN model
%we encode raw visual pixels $1024 \times 1024$ into discrete tokens $64 \times 64$ by VQGAN
with a compression ratio of 16. 
%During training, images are cropped into $1024 \times 1024$ and videos are cut into $1024 \times 1024 \times 5$ with $5$Fps, then, they will be encoded into discrete tokens using the VQGAN model with a compression rate of $16$ and a codebook of $16384$. In Sec.~\ref{sec:rendermodel}, the rendering size of the three models is $256 \times 256$. 
%In Sec.~\ref{sec:renderingstrategy}, based on the nearby sparsity, we set $(e^{h},e^{w},e^{f})=(2,2,0)$ for images and $(e^{h},e^{w},e^{f})=(1,1,3)$ for videos. 
In the second stage, we train the model using Adam optimizer~\cite{kingmaAdamMethodStochastic2014} with a learning rate of $1 \times 10^{-4}$, a batch size of $256$, and warm-up $5$\% of total $50$ epochs. We train four models on four datasets and inference on five tasks. More settings for different tasks can be found in Tab.~\ref{tab:implementation_details}. 
%by default refers to the model of text-to-image synthesis. 
%Texts will be encoded into the tensors of $77 \times 512$ size by pretrained text encoder of CLIP, and they are fed into cross-attention as key and value to interact with visual features. We use three data augmentations on the images including RandomResizedCrop, RandomHorizontalFlip and ColorJitter. The image is only spatially expanded, so we choose $(e^{h},e^{w},e^{f})=(2,2,0)$ expandsion size for image synthesis. NUWA-Infinity refers to the model of image-to-video synthesis. Since it mainly expands the videos in the temporal axis, we choose a smaller spatial receptive field $(e^{h},e^{w})=(1,1)$ but a larger temporal receptive field $e^{f}=3$. Each $1024 \times 1024 \times 5$ clip  is cropped from video with 5fps, and they also apply the three data augmentations mentioned above.  \par
%For training, we employ an Adam optimizer for 50 epochs using 5\% of linear warm-up to a peak learning rate of 1e-4 and a linear decay learning rate scheduler. For inference, we use different sampling strategies for each rendering model. AR models use top-k of 768, NAR models use gumbel sampling and P-NAR models use gumbel sampling with temperature annealing from 4.5 to 1.0.

\subsection{Metrics}

\begin{itemize}[leftmargin=*]

\item \textbf{FID/Block-FID}. Fréchet Inception Distance (FID)~\cite{heuselGansTrainedTwo2017} is used to calculate the quality of the generated image. In this work, we also propose Block-FID, which splits a large image into several blocks and calculates the average Fréchet Inception Distance of all blocks.

% into blocks to calculate Fréchet Inception Distance. It avoids the downsampling caused by resizing images and works for infinite-size synthesis. {\color{red} Since rendering windows and blocks ($256 \times 256$ in our experiments) may overlap, when measuring our model we will generate larger images, and move half of the block size.}
\item \textbf{IS}. Inception Score (IS)~\cite{salimansImprovedTechniquesTraining2016} is a common metric to calculate the diversity of the generated results. A higher IS indicates the model generates more diversified results.
\item \textbf{FVD}. Fréchet Video Distance (FVD) \cite{unterthinerAccurateGenerativeModels2018} is widely used to calculate the quality of generated videos. It measures the distance between the ground-truth video and the generated video. A lower FVD denotes a higher similarity.
\item \textbf{CLIP-SIM}. Recently, the CLIP Similarity Score (CLIP-SIM)~\cite{radfordLearningTransferableVisual2021} is used to measure the  semantic consistency between images and text. We use the CLIP model to calculate the similarity score between the generated image and given text to judge its semantic consistency.
%CLIP Similarity Score (CLIP-SIM)~\cite{radfordLearningTransferableVisual2021} is widely used to calculate the semantic consistency between image and text.
\end{itemize}

\subsection{Main Results}
%\vspace{-2mm}
\subsubsection{Unconditional Image Generation\textsuperscript{\textbf{HD}}}
\vspace{-2mm}

Unconditional Image Generation\textsuperscript{\textbf{HD}} aims to generate images without conditions. Fig.~\ref{fig:uig1} and \ref{fig:uig2} show a single long image generated by the model trained on the RQF dataset for Unconditional Image Generation\textsuperscript{\textbf{HD}}. The generated image has an extremely high resolution of $38912\times 2048$. To better fit the page, we %drop a width of 2 pixels on the right border and we 
split the complete image into 6 splits, each having a resolution of $6485\times 2048$. The generated results demonstrate the following abilities of NUWA-Infinity: 
\begin{itemize}[leftmargin=*]

\item \textbf{Infinity ability}.  NUWA-Infinity can generate visual content with large size. Although trained on $2048 \times 2048$ patches as illustrated in Tab.~\ref{tab:implementation_details}, NUWA-Infinity can generate images 19 times longer than each training instance. This is attributed to our proposed NCP module, which makes the computation grow linearly with the output size, as only a small number of previously generated patches in the context are used during inference. 

\item \textbf{Creation ability}. By comparing the generated results with the original painting, we find that NUWA-Infinity has a strong creative ability. For example, for the gate wall of the second split in Fig.~\ref{fig:uig1}, it is a composition of multiple walls in different directions. Also, for the first split in Fig.~\ref{fig:uig2}, NUWA-Infinity generates a lot of pedestrians and houses. Many different people walk together, which leads to a crowded situation in this split. Note that these scenes do not appear in the original painting, and they are fully created by NUWA-Infinity.

%%%%adjust the two Qingming
\begin{figure}[h]
%  \vspace*{-2.3cm}%-2.3
    \makebox[\linewidth]{
        \includegraphics[width=1.5\linewidth]{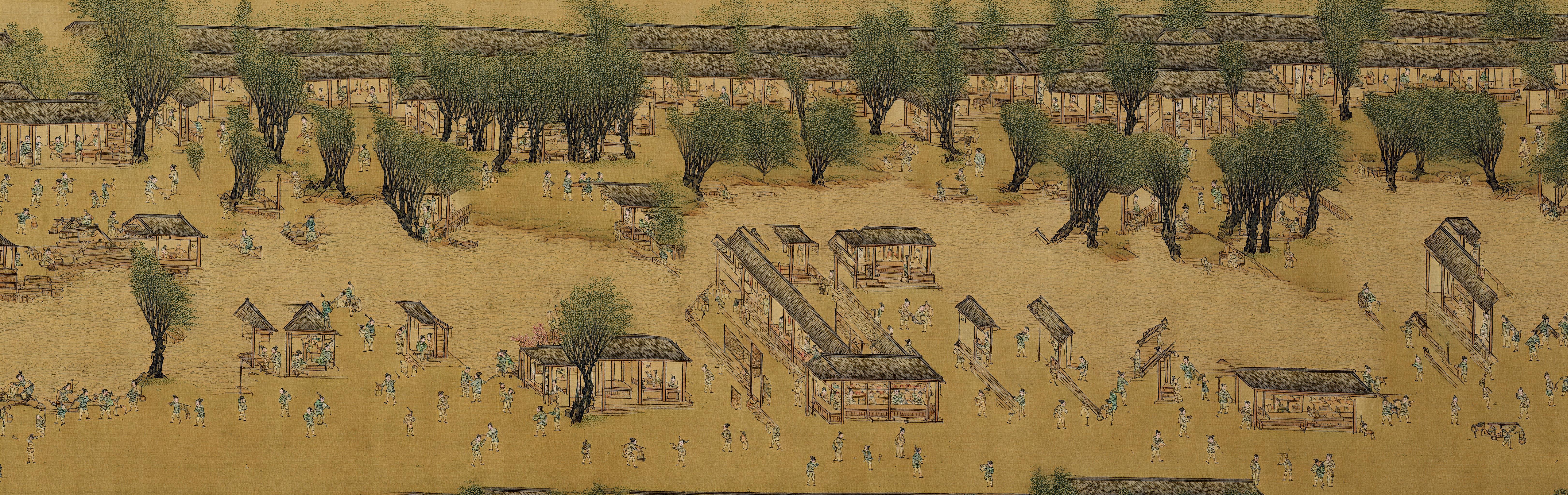}
    }
    \makebox[\linewidth]{
        \includegraphics[width=1.5\linewidth]{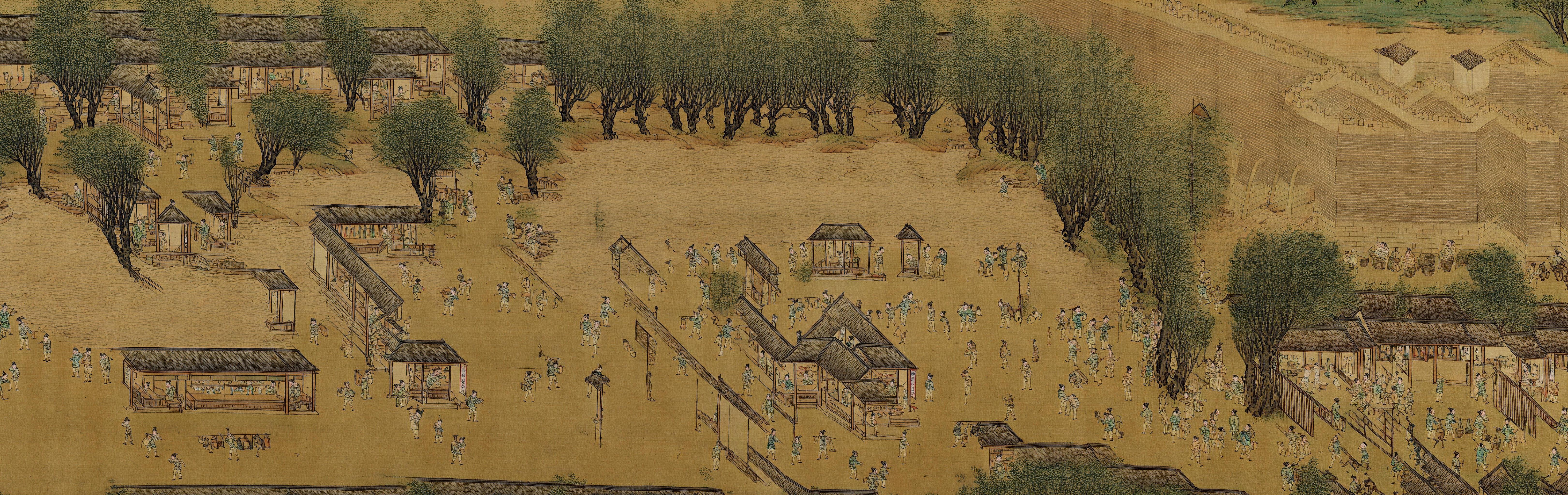}
    }
    \caption{Part I of a huge image ($38912\times 2048$) synthesized in Unconditional Image Generation\textsuperscript{\textbf{HD}} task on RQF dataset. Splits are connectable by row, each has a resolution of $6485\times 2048$.}
    \label{fig:uig1}
\end{figure}
% \vspace{-1cm}

\begin{figure}[p]
 \vspace*{-2.3cm}
    \makebox[\linewidth]{
        \includegraphics[width=1.5\linewidth]{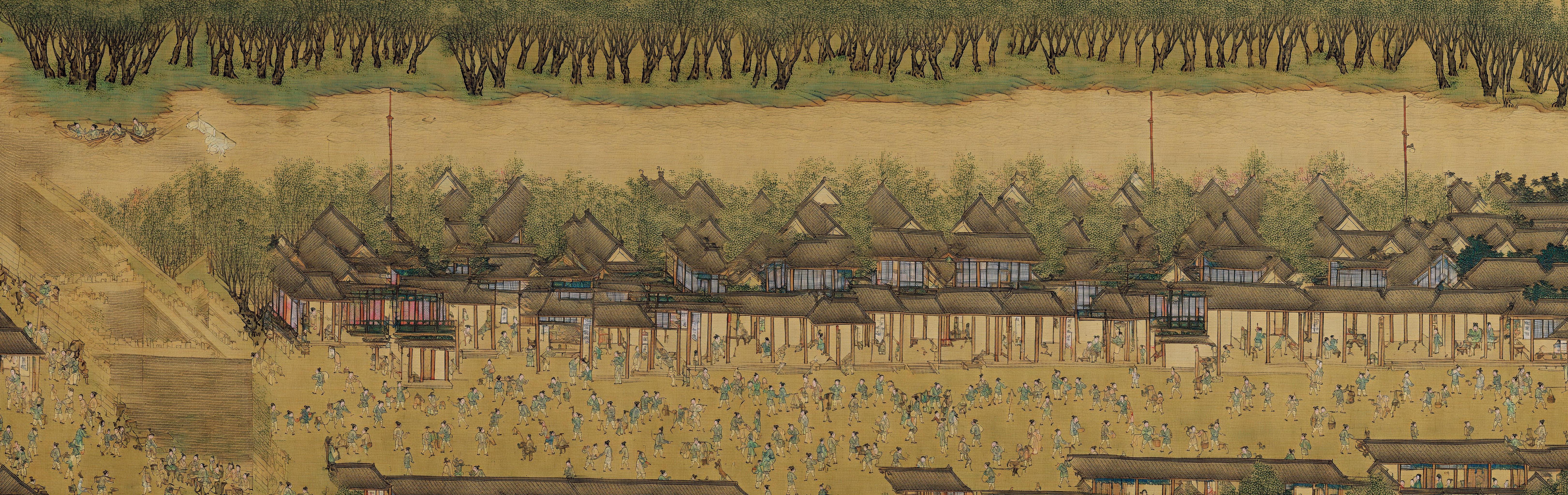}
    }
    \makebox[\linewidth]{
        \includegraphics[width=1.5\linewidth]{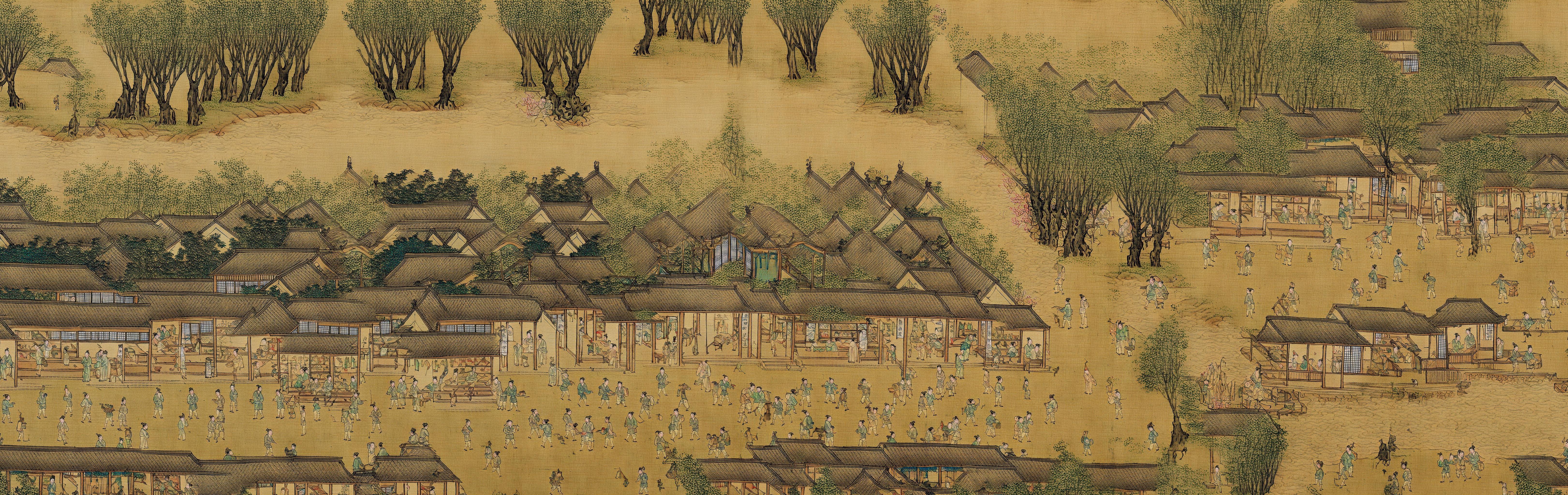}
    }
        \makebox[\linewidth]{
        \includegraphics[width=1.5\linewidth]{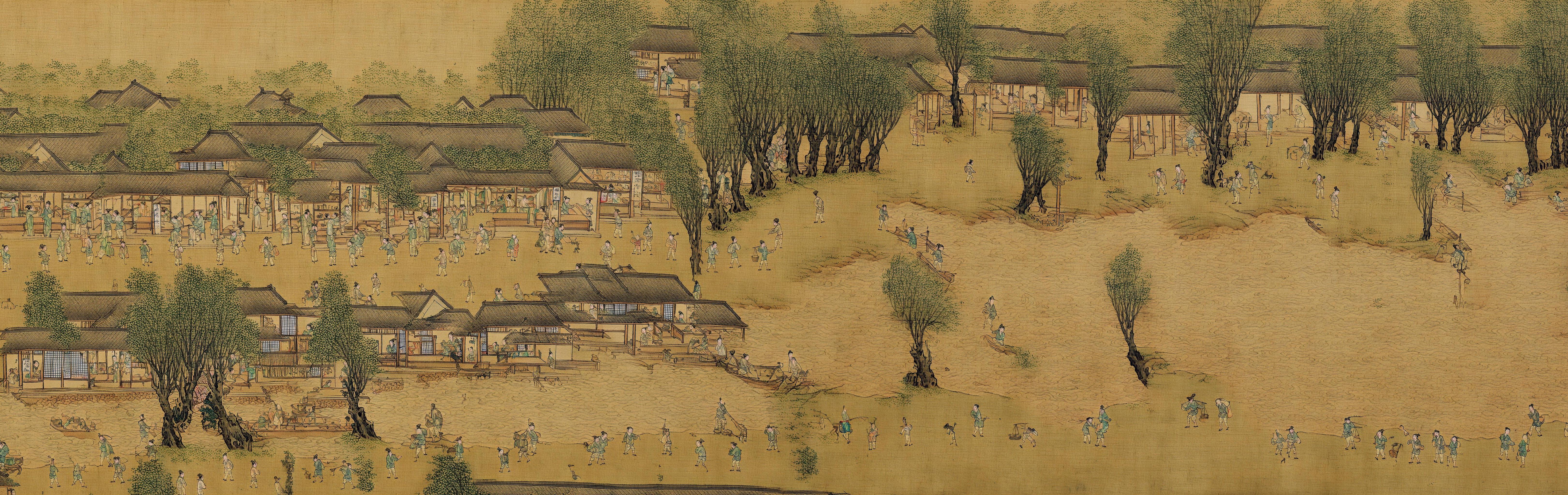}
    }
        \makebox[\linewidth]{
        \includegraphics[width=1.5\linewidth]{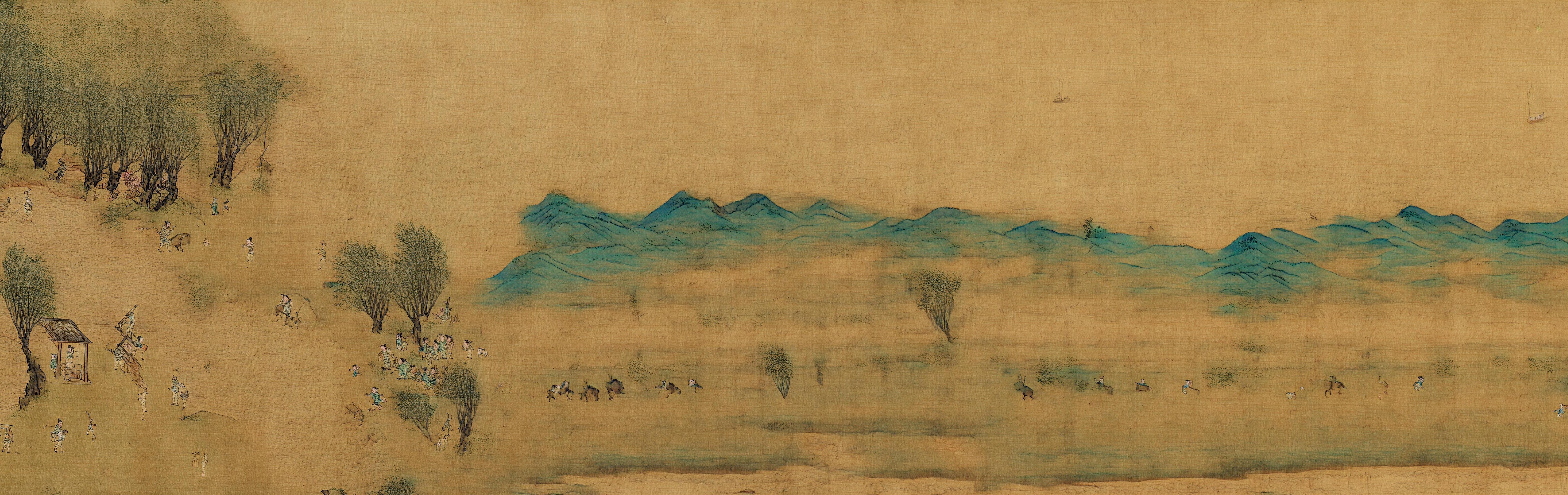}
    }
    % \vspace{-4mm}
    \caption{Part II of the huge image ($38912\times 2048$) illustrated in Fig.~\ref{fig:uig1}.}
    \label{fig:uig2}
\end{figure}

\item \textbf{Local details and global consistency}. The generated results also show decent local details and global consistency. In NUWA-Infinity, the local autoregression generates visual tokens one by one, thus the human faces, human gestures, leaves of trees, roof tiles, and many other details are clearly painted. The global autoregression generates patches one-by-one. As shown in Fig.~\ref{fig:uig2}, the human figures gradually dwindle, and then the picture transitions to the mountains. The smooth transitions make the picture look natural globally even though it is extremely long.

\end{itemize}

\subsubsection{Text-to-Image\textsuperscript{\textbf{HD}}}
\begin{figure}[h]%hbtp
% \begin{figure}[H]
% \begin{figure}[p]
%  \vspace{-1cm}
    % \centering
    \makebox[\linewidth]{
    \includegraphics[width=1.1\linewidth]{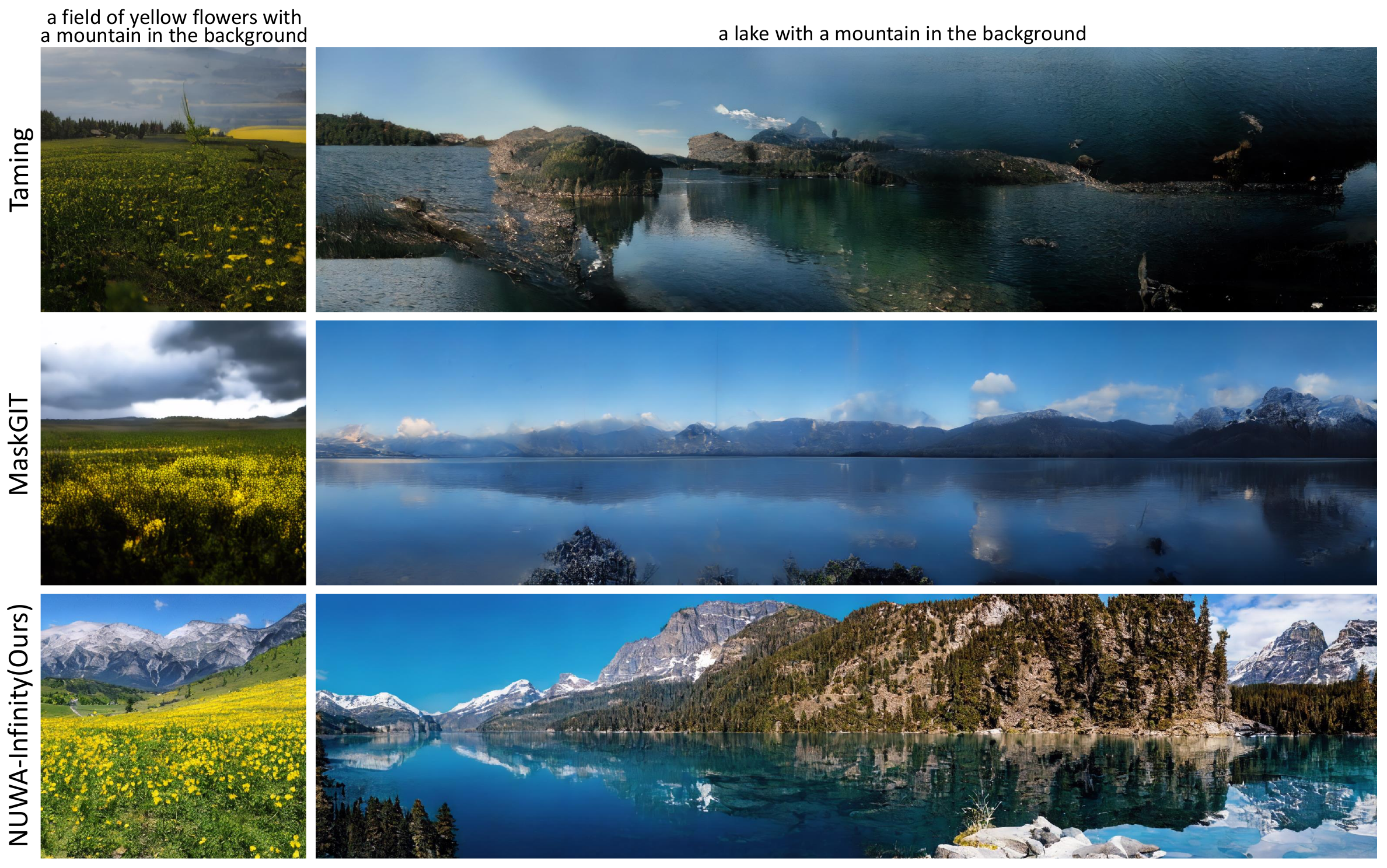}
    }
    \caption{Samples on LHQC dataset for Text-to-Image\textsuperscript{\textbf{HD}}. Left: $1024\times 1024$, Right: $1024\times 4096$. For a fair comparison, we did not provide input sketch for Taming Transformer.}
    \label{fig:vis_t2i}
    	\vspace{-3mm}
\end{figure}

Text-to-Image\textsuperscript{\textbf{HD}} aims to generate an image based on the input text. For a fair comparison, all models are trained from scratch on the LHQC dataset. As shown in Tab.~\ref{tab:t2i_task}, when generating images with $1024 \times 1024$ resolution, NUWA-Infinity outperforms AR-based models Taming Transformer~\cite{esserTamingTransformersHighResolution2021} and Mask-Predict model MaskGIT~\cite{changMaskGITMaskedGenerative2022}, in visual quality (Block-FID), semantic consistency (CLIP-SIM), and diversity (IS). When generating images of size  $4096\times 1024$ which is 4 times as long as the training images, the performance of MaskGIT \cite{changMaskGITMaskedGenerative2022} decreases rapidly but NUWA-Infinity still maintains excellent visual quality with Block-FID of $15.65$. As also shown in Fig.~\ref{fig:vis_t2i}, NUWA-Infinity generates significantly better results, and the reflection of the hill can be clearly seen. Note that we did not compare with DALL·E 2~\cite{rameshHierarchicalTextConditionalImage2022}, Imagen \cite{sahariaPhotorealisticTexttoImageDiffusion2022}, or Parti~\cite{yuScalingAutoregressiveModels2022} directly because of three reasons: ($i$) all of them do not support arbitrary-large visual synthesis (i.e, generating images with $4096\times 1024$ resolution);  ($ii$) they did not make their pre-trained models public; and ($iii$) NUWA-Infinity focuses on enabling infinite visual synthesis and is not pre-trained on large-scale datasets, thus it is hard to make a direct comparison.

% \begin{figure}[hbtp]
%  \vspace*{-2cm}
%     % \centering
%     \makebox[\linewidth]{
%     \includegraphics[width=1.3\linewidth]{Figure/t2i_compare.pdf}
%     }
%     \caption{High Resolution text-to-image, the left is $1024\times 1024$, the right is $1024\times 4096$.}
%     \label{fig:vis_t2i}
%     	\vspace{-2mm}
% \end{figure}

%  image an video table
\begin{table}[htbp]
\small
\centering
\setlength{\tabcolsep}{14pt}
\begin{tabular}{lcccc}
 \toprule
Method    & Block-FID↓    & Block-FID($\times 4$)↓ & IS↑           & CLIP-SIM↑       \\    
\midrule
Taming \cite{esserTamingTransformersHighResolution2021}    & 38.89         & 46.37          & 4.58          & 0.2662          \\
MaskGIT \cite{changMaskGITMaskedGenerative2022}   & 24.33         & 45.76          & 4.61          & 0.2754          \\
NUWA-Infinity & \textbf{9.71} & \textbf{15.65} & \textbf{4.98} & \textbf{0.2807} \\
 \bottomrule
\end{tabular}
\vspace{-2mm}
\caption{Comparisons on LHQC dataset for Text-to-Image\textsuperscript{\textbf{HD}} task.}
\label{tab:t2i_task}
\end{table}

% \begin{table}[htbp]
% \centering
% \begin{tabular}{lc}
% \toprule
% Method         & FVD↓           \\
% \midrule
% NUWA-Infinity (NAR)   & 368.16          \\
% NUWA-Infinity (P-NAR) & 165.39          \\
% NUWA-Infinity (AR)    & \textbf{62.57} \\
% \bottomrule
% \end{tabular}
% \caption{Video prediction on LHQ-V, the size of sample is $1024\times 1024\times 8$.}
% \label{tab:i2v_task}
% \end{table}

\subsubsection{Image Animation\textsuperscript{\textbf{HD}}}
\begin{figure}[h]
    \centering
    \includegraphics[width=1\textwidth]{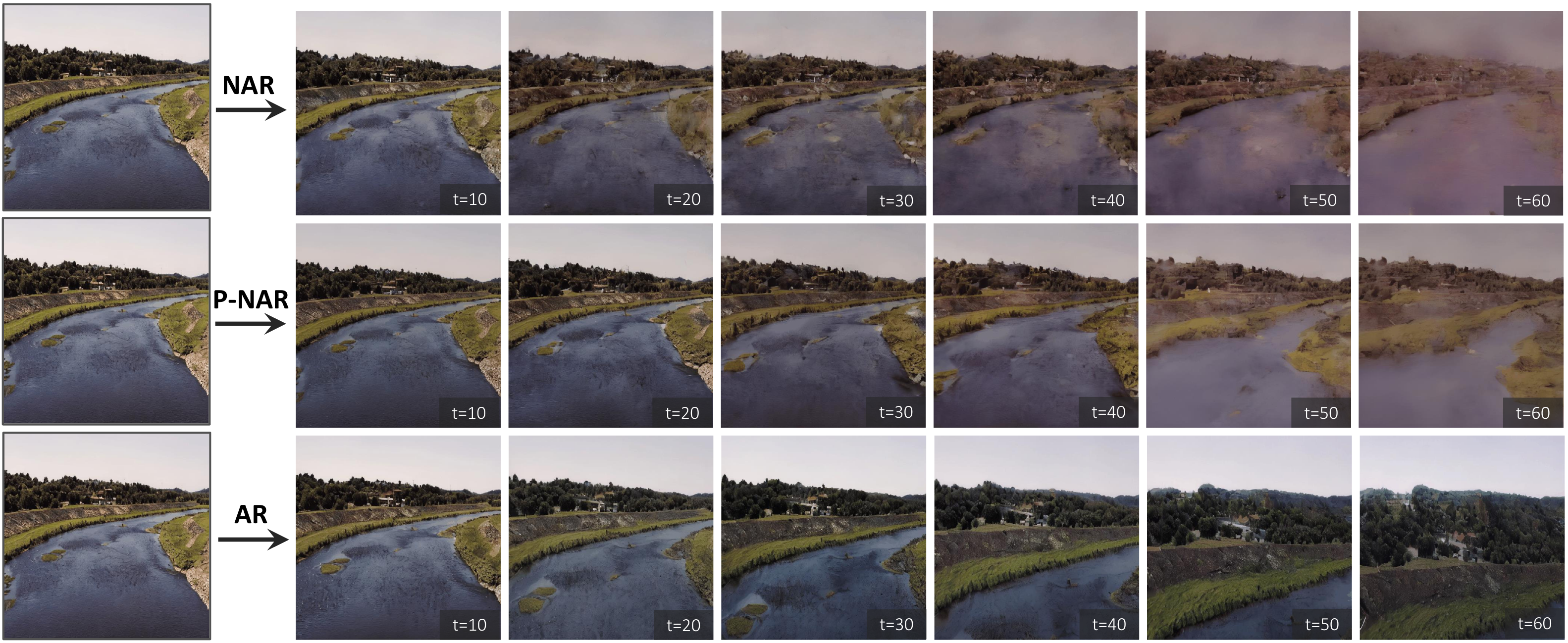}
    \caption{Samples on LHQ-V dataset for Image Animation\textsuperscript{\textbf{HD}} task. We compare local NAR, P-NAR, and AR by generating $60$ frames with a high resolution of $1024\times 1024$.}
    \label{fig:vis_i2v}
	\vspace{-3mm}
\end{figure}

% \begin{wraptable}[8]{r}{0.5\textwidth}
% \renewcommand{\arraystretch}{0.8}
% \setlength{\tabcolsep}{1.5pt}
\begin{table}
\centering
\begin{tabular}{lc}
\toprule
Method        & FVD↓           \\
\midrule
NUWA-Infinity (NAR)   &368.16          \\
NUWA-Infinity (P-NAR) & 165.39          \\
NUWA-Infinity (AR) & \textbf{62.57} \\
\bottomrule
\end{tabular}
\caption{Comparisons on LHQ-V dataset for Image Animation\textsuperscript{\textbf{HD}} task.}
\label{tab:i2v_task}
\end{table}
% \end{wraptable}
Image Animation\textsuperscript{\textbf{HD}} aims to generate a video based on an input image. We build two baseline models as a comparison shown in Tab.~\ref{tab:i2v_task}. All three models are globally autoregressive, but they use different local generative methods (shown in round brackets). NUWA-Infinity (AR) is our default autoregressive over autoregressive model with a local autoregressive mechanism to generate a patch. NUWA-Infinity (NAR) can be viewed as an autoregressive over non-autoregressive model, as the tokens inside a patch are generated locally in parallel instead of autoregressively. NUWA-Infinity (P-NAR) can be viewed as an autoregressive over progressive non-autoregressive model, as locally, the tokens inside a patch are generated by a Mask-Predict method~\cite{changMaskGITMaskedGenerative2022} introduced in Sec.~\ref{sec:relatedwork}. NUWA-Infinity achieves significantly better performance than baselines, with an FVD score of 62.57. Fig.~\ref{fig:vis_i2v} provides a qualitative comparison between the generated $60$-frame videos from the same input image of $1024\times 1024$ resolution. We find that NUWA-Infinity with a default local AR mechanism can generate more realistic images. However, we do find a speed-performance trade-off as AR generation takes more time compared with NAR and P-NAR. We will discuss it in Sec.~\ref{sec:ablvd}.

% figure image-to-video

\subsubsection{Image Outpainting\textsuperscript{\textbf{HD}}}
Image Outpainting\textsuperscript{\textbf{HD}} aims to generate an out-painted image based on the input image. We do not train NUWA-Infinity for the Image Outpainting\textsuperscript{\textbf{HD}} task, but used the model trained for Text-to-Image\textsuperscript{\textbf{HD}} directly. To evaluate the model's ability of outpainting in all directions, we set up four settings: Right Extend $\Rightarrow$, Left Extend $\Leftarrow $, Down Extend $\Downarrow$ and Up Extend $\Uparrow$. For example, in the Right Extend $\Rightarrow$ setting, we input the image with the left half in the LHQC dataset and ask the model to predict the right half. Since for the output image in this task, half is ground-truth and half is extended, we calculate Block-FID between the extended area and the same area in the ground-truth test set of LHQC. As shown in Tab.~\ref{tab:ie}, NUWA-Infinity significantly outperforms Taming~\cite{esserTamingTransformersHighResolution2021} and MaskGIT~\cite{changMaskGITMaskedGenerative2022} by a large margin in all four directions. Since the model trained by Text-to-Image\textsuperscript{\textbf{HD}} also supports a text prompt, we also try to outpaint the image by adding text controlling, and we find better performance on Down Extend $\Downarrow$ and Up Extend $\Uparrow$, while similar performance on Right Extend $\Rightarrow$ and Left Extend $\Leftarrow $ compared with one without text. We hypothesize that it is because the upper or lower half of the image contains less information, while the left or right half has more visual semantic hints. Note that for a fair comparison, Taming and MaskGIT use text prompt by default.

\begin{figure}[t!]
    \centering
    \includegraphics[width=1\textwidth]{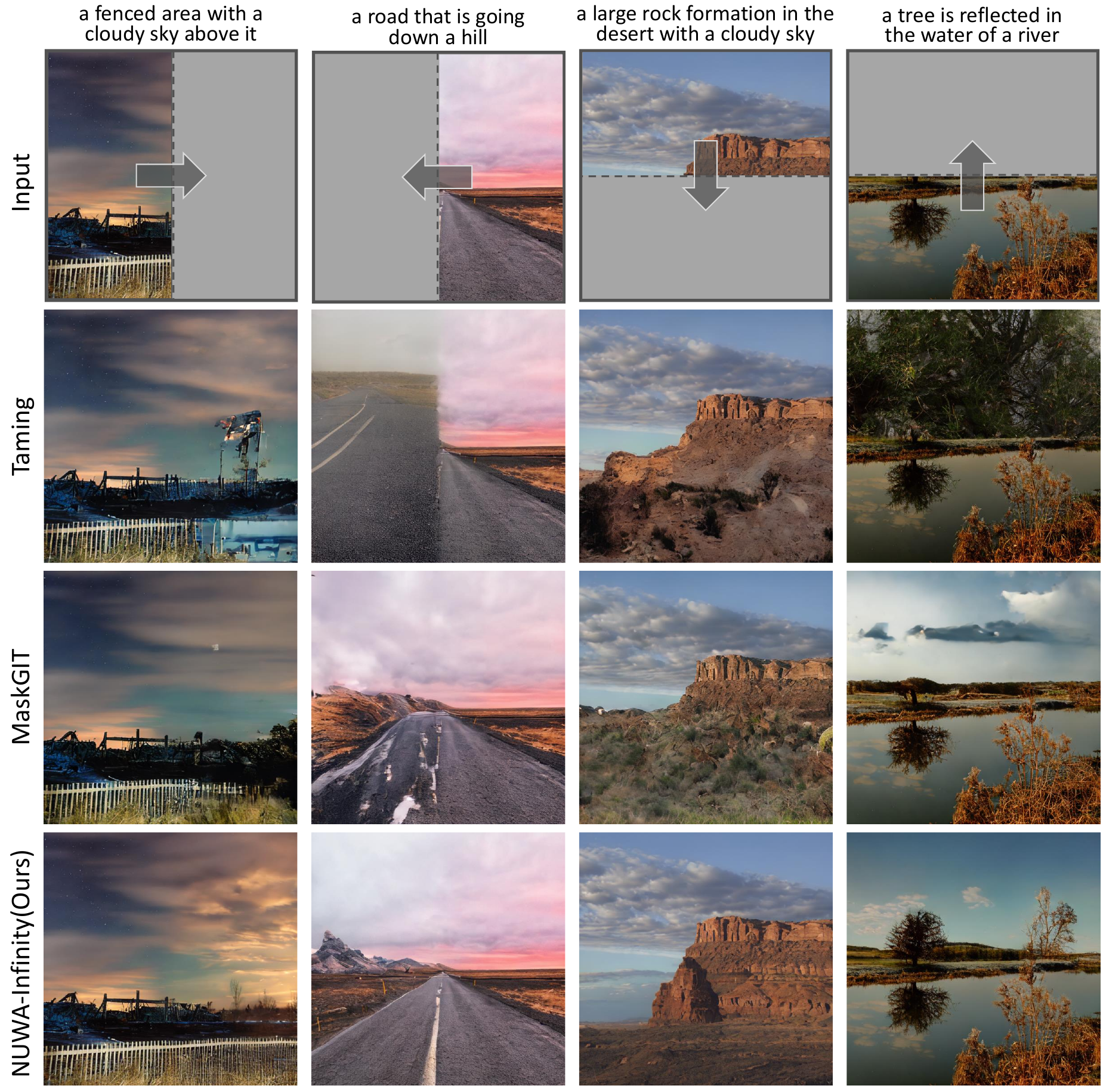}
    \caption{Samples for Image Outpainting\textsuperscript{\textbf{HD}} task on LHQC dataset. The input image has a resolution of half of $1024\times 1024$.}
    \vspace{-4mm}
    \label{fig:vis_ie}
\end{figure}

In Fig.~\ref{fig:vis_ie}, we provide four input images illustrating four directions. Taming Transformer only successfully generates the input image in the third column, as it predicts the lower half of the hill based on the upper half. This is because Taming Transformer only trains token-by-token in $\zeta-order$, and this order only fits the down extension. MaskGIT successfully predicts another half in all directions benefited by its bidirectional masked language model. Compared with MaskGIT, NUWA-Infinity can generate more realistic results. For example, in the fourth column, when inputting the reflection of a tree in the lake, NUWA-Infinity successfully generates the most consistent tree on the shore.
% figure image extension
% table image extension
\begin{table}[htbp]
\small
\begin{tabular}{lcccc}
\toprule
\multirow{2}{*}{Method} & \multicolumn{4}{c}{Block-FID↓}                                \\ \cmidrule{2-5} 
                        & Right Extend $\Rightarrow$   & Left Extend $\Leftarrow $   & Down Extend $\Downarrow$  & Up Extend $\Uparrow$    \\ 
                        \midrule
Taming \cite{esserTamingTransformersHighResolution2021}                  & 22.53         & N/A           & 26.38         & N/A           \\
MaskGIT \cite{changMaskGITMaskedGenerative2022}                 & 14.68         & 14.81         & 25.57         & 25.38         \\
NUWA-Infinity w/o text      & \textbf{6.43} & \textbf{6.71} & 11.47         & 8.03          \\
NUWA-Infinity       & 6.45          & 6.72          & \textbf{9.84} & \textbf{7.43} \\
\bottomrule
\end{tabular}
\caption{Comparisons on LHQC dataset for Image Outpainting\textsuperscript{\textbf{HD}} task.}
\label{tab:ie}
\end{table}

% \subsubsection{Text-to-Video\textsuperscript{\textbf{HD}}}
% \vspace{-2mm}
% Text-to-Video\textsuperscript{\textbf{HD}} aims to generate high-resolution videos given the input texts. As shown in Fig.~\ref{fig:pos}, NUWA-Infinity can generate videos considering both objects and motions, and distinguish the motion ``up the hill'' from ``down the hill''. Based on this ability, Fig.~\ref{fig:peppapig} shows a simple video story generated by four sentences. We believe this technology could help cartoon design in the future.

% \begin{figure}[t]
%     \centering
%     \includegraphics[width=1.0\textwidth]{Figure/Pig.pdf}
%     \caption{Samples for Text-to-Video\textsuperscript{\textbf{HD}} task on PeppaPig dataset. The generated video has a resolution of $1024\times 1024$ with 8 frames.}
%     \vspace{-4mm}
%     \label{fig:peppapig}
    
% \end{figure}

\subsection{Ablation Studies}
\vspace{-2mm}
We conduct detailed ablations on three components of our model, ADC, NCP, and Vision Decoder. 

\subsubsection{Ablation Study on ADC}
\vspace{-2mm}
\paragraph{Patch size of ADC.Split}
\begin{wrapfigure}[11]{r}{0.5\textwidth}
\vspace{-0.17in}
    \centering
    \includegraphics[scale=0.45]{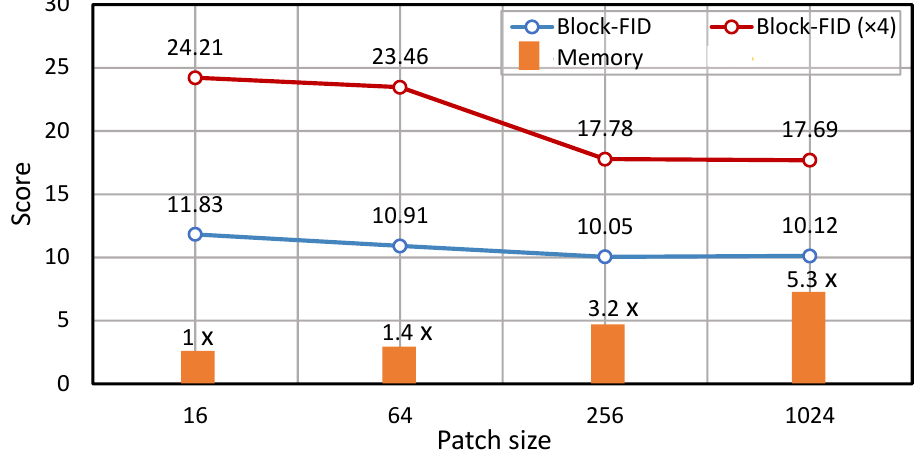}
    \vspace{-1.5em}
    \caption{Impact of patch size.}
    \label{fig:adcsplit}
\end{wrapfigure}
As introduced in Sec.~\ref{sec:train}, each patch representation has a size of $p_n\in \mathbb{R}^{M\times d}$. The horizontal axis of Fig.~\ref{fig:adcsplit} shows different patch sizes $M$, and the vertical axis shows FID scores for Text-to-Image\textsuperscript{\textbf{HD}}. The blue line shows the FID score of generating image resolution of $1024\times 1024$. The red line shows the Block-FID score of generating image resolution of $1024\times 4096$. A smaller patch size harms the FID score of the generated image, but a larger patch size requires larger GPU memory. When patch size $M=256$, a good balance between performance and memory can be achieved.

\paragraph{Feed position of ADC.Emb} 
\begin{wraptable}[6]{r}{0.42\textwidth}
\vspace{-0.25in}
\centering
\vskip 0.1in
\setlength{\tabcolsep}{2.5pt}
\begin{tabular}{lcc}
    \toprule
    RPE  & Block-FID↓ & Block-FID(×4) ↓ \\ \hline
    %\rowcolor[HTML]{E6E6E6} 
    Pre  & \textbf{10.05}      & \textbf{17.78}         \\
    Post & 10.47      & 18.89  \\  
    \bottomrule
    \end{tabular}
\caption{Impact of feed position.}
 \label{tab:adcemb}
\end{wraptable}
As introduced in Sec.~\ref{sec:adc}, ADC dynamically provides relative positional embeddings during training and inference stages. In Eq.~\ref{eq:qkv}, the relative positional embedding $e_n$ is added to the key of self-attention. We call it pre-feeding, as the positional information is fed before the computation of attention. This is different from traditional post-feeding in Transformers, where the positional information is fed after the computation of attention. Tab.~\ref{tab:adcemb} shows that pre-feeding brings better performance than post-feeding. This is because pre-feeding controls which contexts to attend to with positional information while post-feeding only adjusts the attention distributions after the attention.

%%%%original 
% \begin{figure}[h]
%       \centering
%       \includegraphics[width=0.5\textwidth]{Figure/ab_rendersize.pdf}
%         \caption{Rendering size.}
%         \label{fig:rendersize}
% \end{figure}
%%%%%%

%%%%%test
% \begin{figure}[h]
% \begin{floatrow}
% \ffigbox{%
%   \includegraphics[width=0.5\textwidth]{Figure/ab_patchsizeV2.pdf}
% }{%
%   \caption{Patch size of ADC.Split}
%     \label{fig:adcsplit}
% }
% \capbtabbox{%
%     \begin{tabular}{lcc}
%     \toprule
%     RPE  & Block-FID↓ & Block-FID(×4) ↓ \\ \hline
%     %\rowcolor[HTML]{E6E6E6} 
%     Pre  & \textbf{10.05}      & \textbf{17.78}         \\
%     Post & 10.47      & 18.89  \\  
%     \bottomrule
%     \end{tabular}
% }{
%   \caption{Feed Position of ADC.Emb.}
%       \label{tab:adcemb}

% }
% \end{floatrow}
% \end{figure}
%%%%%test

% %%%%****original RPE table
% \begin{table}
% \centering
% \begin{tabular}{lcc}
% \toprule
% RPE  & Block-FID↓ & Block-FID(×4) ↓ \\ \hline
% %\rowcolor[HTML]{E6E6E6} 
% Pre  & \textbf{10.05}      & \textbf{17.78}         \\
% Post & 10.47      & 18.89  \\  
% \bottomrule
% \end{tabular}
% \caption{Patch size of ADC.Split.}
% \label{tab:rpe}
% \end{table}
% %%%%%%%%%original RPE table

\subsubsection{Ablation Study on NCP}

\paragraph{Caches in NCP}
\begin{wraptable}[7]{r}{0.46\textwidth}
\vspace{-0.16in}
\centering
\setlength{\tabcolsep}{2pt}
\begin{tabular}{lcc}
\toprule
Context      & Block-FID↓ & Block-FID(×4) ↓ \\ 
\midrule
w/o caches & 15.80       & 38.32           \\
w/ caches  & \textbf{10.21}      & \textbf{23.62} \\
\bottomrule
\end{tabular}
\caption{Ablation results in caches in NCP.}
\label{tab:transfer}
\end{wraptable}

% \begin{wraptable}
% \vspace{-0.25in}
% % \renewcommand{\arraystretch}{0.7}
% \centering
% \vskip 0.1in
% \setlength{\tabcolsep}{2.5pt}
% \begin{tabular}{lcc}
%     \toprule
%     RPE  & Block-FID↓ & Block-FID(×4) ↓ \\ \hline
%     %\rowcolor[HTML]{E6E6E6} 
%     Pre  & \textbf{10.05}      & \textbf{17.78}         \\
%     Post & 10.47      & 18.89  \\  
%     \bottomrule
%     \end{tabular}
% \caption{Impact of feed position.}
%  \label{tab:adcemb}
% \end{wraptable}
As introduced in Sec.~\ref{sec:ncp}, the Add operation saves multi-layer hidden states of previously generated patches as ``caches''. 
%One crucial design is that the caches %are hidden states of previous patches. 
This allows information transmission between patches during training and inference. In other words, even though distant patches are removed from NCP, their information can be still captured in the hidden states of the nearby patches in NCP. 
%To verify the effectiveness of this design, we cut off information transmission by simply inferring the multi-layer hidden states without NCP, as shown in  Tab.~\ref{tab:transfer}.
To verify the effectiveness of this design, we train another model without using the caches in NCP and show its result of Text-to-Image\textsuperscript{\textbf{HD}} in Tab.~\ref{tab:transfer}.
The NCP design with information transmission significantly outperforms the one without information transmission.

% \begin{figure}
% \centering
% \begin{minipage}[b][0.5\linewidth]
% \begin{tabular}{lcc}
% \toprule
% %\hline
% Context      & Block-FID↓ & Block-FID(×4) ↓ \\ \hline
% w/o transfer & 15.8       & 38.32           \\
% %\rowcolor[HTML]{E6E6E6} 
% w/ transfer  & \textbf{10.21}      & \textbf{23.62} \\
% \bottomrule
% %\hline
% \label{tab:transfer}
% \end{tabular}
% \end{minipage}
% \caption{Context transfer.}
% \end{figure}

\begin{figure}
\centering
\begin{subfigure}[t]{0.44\textwidth}
    \centering
    \includegraphics[width=\textwidth]{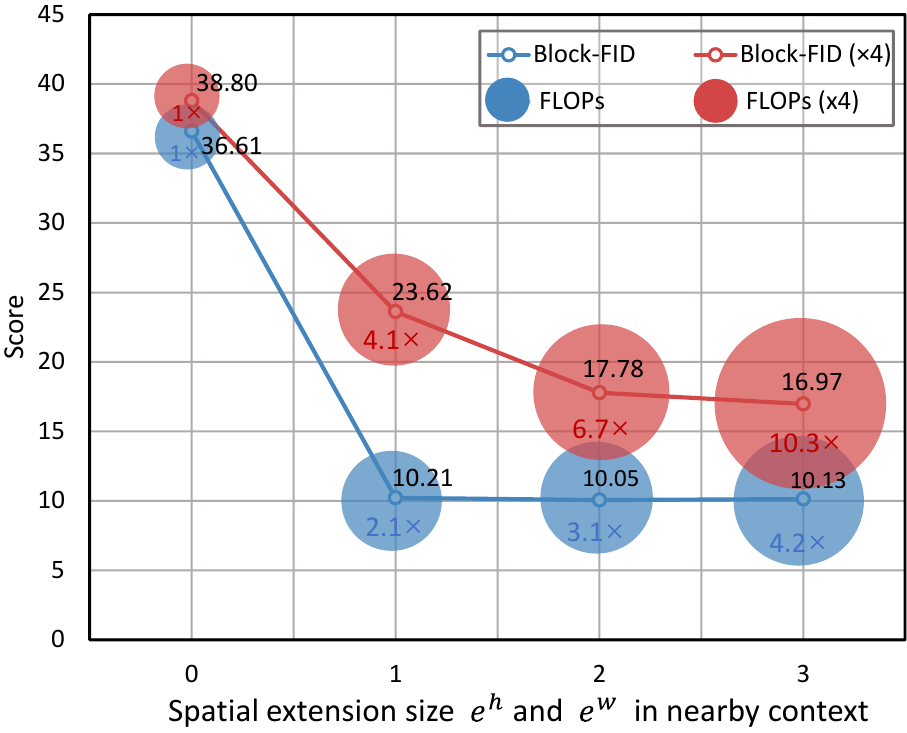}
    % \centerline{(a) Spatial extension}
    \caption{Spatial extension.}
    \label{fig:spatialextend}
\end{subfigure}
\begin{subfigure}[t]{0.44\textwidth}
    \centering
    \includegraphics[width=\textwidth]{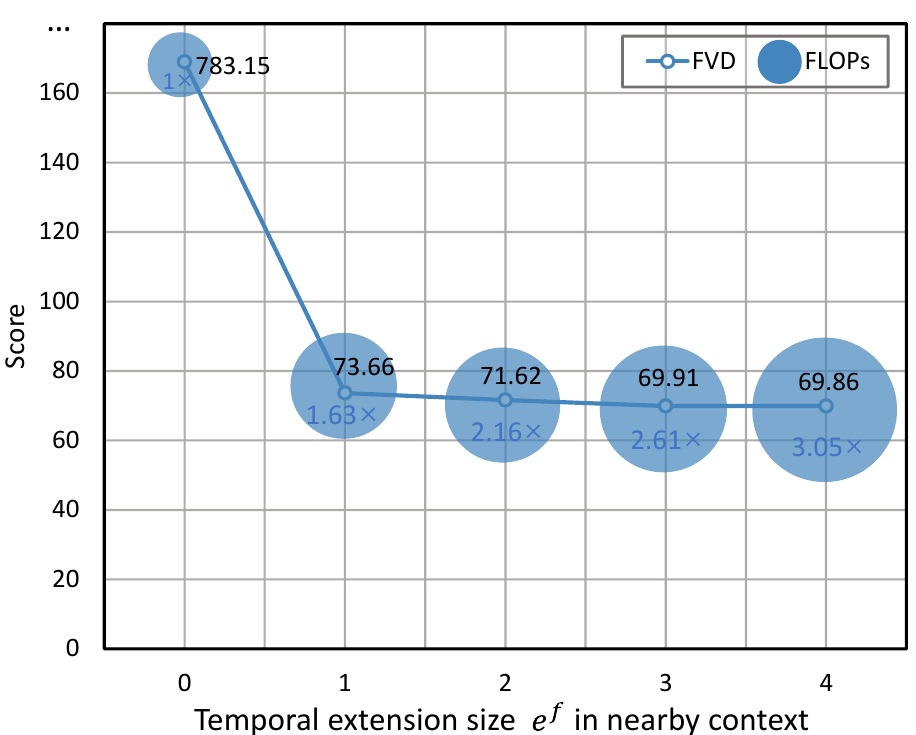}
    % \centerline{(b) Temporal extension}
    \caption{Temporal extension.}
    \label{fig:temporalextend}
\end{subfigure}
\vspace{1.5em}
\caption{Ablation results on extents in NCP.}

\vspace{-2em}
\end{figure}

% \begin{table}[h]
% \centering
% \begin{minipage}{0.9\linewidth}
% \centering
% \renewcommand{\arraystretch}{0.6}
% \setlength{\tabcolsep}{20pt}
% \begin{tabular}{lcc}
% \toprule
% Context      & Block-FID↓ & Block-FID(×4) ↓ \\ 
% \midrule
% w/o transfer & 15.80       & 38.32           \\
% w/ transfer  & \textbf{10.21}      & \textbf{23.62} \\
% \bottomrule
% \end{tabular}
% \label{tab:transfer}
% \vspace{0.3em}
% \centerline{(a) Context transfer}

% \end{minipage}
% \vspace{1em}
% % \hspace{0.1in}
% \begin{minipage}{\linewidth}
% \centering
% \begin{subfigure}[t]{0.44\textwidth}
%     \centering
%     \includegraphics[width=\textwidth]{Figure/ab_spatial.pdf}
%     \centerline{(b) Spatial extension}
%     % \caption{Spatial extension.}
%     \label{fig:spatialextend}
% \end{subfigure}
% \begin{subfigure}[t]{0.44\textwidth}
%     \centering
%     \includegraphics[width=\textwidth]{Figure/ab_temporal.pdf}
%     \centerline{(c) Temporal extension}
%     % \caption{Temporal extension.}
%     \label{fig:temporalextend}
% \end{subfigure}
% \caption{Ablation results on nearby expansion size, (a) and (b) are based on text-to-image synthesis, (c) is based on video prediction with $8$ frames. Note that they all use the base model with layers of $16$ and dim of $768$.}

% \vspace{-2em}
% \end{minipage}
% \end{table}

\paragraph{Extents in NCP}
As introduced in Sec.~\ref{sec:ncp}, NCP saves caches in a 3D extent $(e^w, e^h, e^f)$. Fig.~\ref{fig:spatialextend} shows the comparisons between different extent size of $(e^w, e^h, e^f)$ for Text-to-Image\textsuperscript{\textbf{HD}}. Although larger extent size brings better Block-FID scores, the performance becomes limited when $e^{h},e^{w}>2$ while computation improves significantly. For videos, Fig.~\ref{fig:temporalextend} shows the impact of temporal extent on Image Animation\textsuperscript{\textbf{HD}}. We find the FVD improvements become limited if the $e^f$ increases to 3. As a result, we choose $(e^w, e^h, e^f)=(2,2,3)$ as our default setting. 

% \begin{figure*}[t!]

%     \begin{subfigure}[t]{0.33\textwidth}
%             \centering
%             \includegraphics[width=\textwidth]{Figure/ab_spatial.pdf}
%             \caption{Spatial extension.}
%             \label{fig:spatialextend}
%     \end{subfigure}
%     \begin{subfigure}[t]{0.33\textwidth}
%             \centering
%             \includegraphics[width=\textwidth]{Figure/ab_temporal.pdf}
%             \caption{Temporal extension.}
%             \label{fig:temporalextend}
%     \end{subfigure}
%     \caption{Ablation results on nearby expansion size, (a) and (b) are based on text-to-image synthesis, (c) is based on video prediction with $8$ frames. Note that they all use the base model with layers of $16$ and dim of $768$.}
%     \vspace{-5mm}
% \end{figure*}

\subsubsection{Ablation Study on Vision Decoder}\label{sec:ablvd}
Tab.~\ref{tab:rendermodel} shows ablations on different vision decoders. NUWA-Infinity is an autoregressive over autoregressive model and it follows the autoregressive (AR) formulation in the vision decoder. We also try another two vision decoders based on a non-autoregressive (NAR) formulation and a progressive autoregressive (P-NAR) formulation. For these two models, the global autoregression is still maintained, while the local autoregression is changed into NAR and P-NAR, respectively. We find AR-based vision decoder achieves the best performance and NAR-based vision decoder achieves the fastest speed.

Tab.~\ref{tab:modelsize} shows two settings of NUWA-Infinity: Base and Large. We find that the base model can also achieve acceptable performance compared with the large model. This is due to the limited training samples in the dataset. In this paper, we focus on the effectiveness of NUWA-Infinity architecture, instead of large-scale pre-training. We will pre-train NUWA-Infinity with more data in the future.

Tab.~\ref{tab:lossregion} shows when to optimize the loss between the predicted patch of vision decoder and the ground-truth patches. During training, as long as a patch is predicted, NUWA-Infinity calculates the patch loss and optimizes it immediately instead of accumulating the gradients of all patches. We simply call this mechanism ``patch loss'' and the accumulated one ``accumulated loss''. We find that patch loss accelerates convergence from $70$ epochs to $50$ epochs and improves the Block-FID score compared with the accumulated loss. This is because patch loss will share optimized parameters between each patch, which helps the model learn large images or videos.

% ablation study big table
\begin{table}[t!] 
\small
\begin{subtable}{1.\textwidth}
\centering 
 \begin{tabular}{lcccc} 
\toprule
Vision Decoder & Block-FID↓ & Block-FID($\times$ 4)↓ & CLIP-SIM↑ & Inference Speed↑ \\ \hline
NUWA-Infinity (NAR)   & 92.34 & 98.67 & 0.2451  & \textbf{95}$\times$ \\
NUWA-Infinity (P-NAR) & 19.86 & 38.59 & 0.2726  & 15$\times$          \\
NUWA-Infinity (AR)    & \textbf{10.05} & \textbf{17.78} & \textbf{0.2753}    & 1$\times$  \\
\bottomrule
\end{tabular}
\caption{Decoder model.}
\label{tab:rendermodel}
\end{subtable}
% \hspace{-0.2em}
\begin{subtable}{.48\linewidth}
\setlength{\tabcolsep}{4.5pt}
% \Leftarrow
\begin{tabular}{lccc}
\toprule
Parameters  & Depth & Dim  & Block-FID↓ \\ \hline
%\rowcolor[HTML]{E6E6E6} 
%%%if Base->B, Large->L
202M (Base)  & 16    & 768  & 10.05      \\
809M (Large) & 24    & 1280 & \textbf{9.71}  \\
\bottomrule
\end{tabular}
\caption{Decoder size.}
\label{tab:modelsize}
\end{subtable}
\hspace{0.6em}
\begin{subtable}{.45\linewidth}
\setlength{\tabcolsep}{1.5pt}
\centering
\begin{tabular}{lcc}
\toprule
Loss  & Block-FID↓ & Convergence epoch↓ \\ \hline
%\rowcolor[HTML]{E6E6E6} 
Patch & \textbf{10.05}         & \textbf{50}                 \\
Accumulated  & 11.62         & 70   \\ 
\bottomrule
\end{tabular}
\caption{Decoder loss.}
\label{tab:lossregion}
\end{subtable}
\caption{Ablation experiments with text-to-image generation on LHQC. We use a base model with $16$ layers and dim of $768$ except (b). } \label{tab:ablation_five}
% \vspace{-5mm}
% The default setting is marked in gray.
\end{table}

\section{Discussions}

\paragraph{Training Data}
The model for infinite visual synthesis requires high-resolution images and videos as the training data. Such high-quality visual data are harder to collect duo to quality and license issues. In the future, we will collect large-scale datasets satisfying the quality and license criteria for the development of this research direction.

\paragraph{Evaluation Metric}
Compared to text generation tasks, such as machine translation and text summarization, visual synthesis is more difficult to evaluate as the number of ground-truths of a model's output could be unlimited. Currently, we follow the traditions (i.e., using FID, FVD, IS, and CLIP-SIM scores) to measure the quality of generated images and videos. In the future, we will explore better evaluation metrics for visual synthesis tasks.

\paragraph{Inference Speed} Autoregressive models can deal with dependencies between generated contents well. However, the training and inference efficiency of this generation mechanism is still a blocking issue for deploying such models for practical usage. In the future, we will explore ways to combine the advantages of autoregressive models and non-autoregressive models (such as the diffusion model) to achieve both generation quality and inference (or training) efficiency.

\paragraph{Pre-trained Version} In this paper, we train NUWA-Infinity for different downstream tasks directly, due to the lack of large-scale high-quality visual data. In the future, we will pre-train the next version of NUWA-Infinity with more collected visual data and report its generalization  capabilities on open-domain inputs.

\section{Conclusion}
NUWA-Infinity is a visual synthesis framework that can be trained to generate high-quality images and videos from the given text or image input. Different from DALL·E, DALL·E 2, Imagen and Parti, an autoregressive over autoregressive mechanism is proposed to support variable-size visual content generation tasks, such as image outpainting, image animation, text-to-image generation, and text-to-video generation. We hope such models help visual content creators save time, cut costs, and increase productivity and creativity.
%In this paper, we propose NUWA-Infinity as a generative model for infinite visual synthesis. An autoregressive over autoregressive generation mechanism is proposed to deal with this variable-size generation task, where a global patch-level autoregressive model considers the dependencies between patches, and a local token-level autoregressive model considers the dependencies between visual tokens within each patch. Meanwhile, A Nearby Context Pool (NCP) and an Arbitrary Direction Controller (ADC) is introduced to cache related patches already generated as the context and decide suitable generation orders for different visual synthesis tasks. 

\section*{Acknowledgements}

We'd like to thank Minheng Ni, Xiaodong Wang, and Bei Li for the figure and table formats of this paper. We'd also like to thank Yu Liu, Jieyu Xiao, Scarlett Li, and Jane Ma for the discussion of potential application scenarios. We'd also like to thank Yang Ou and Bella Guo for the design of the homepage, and Tiantian Xue and Daisy Hou for the implementation of the homepage. We'd also like to thank Ting Song, Yan Xia, and Shiyou Ren for the help with the dataset construction. We'd also like to thank Yan Fan and Quanlu Zhang for their system support.

\bibliographystyle{plain}
\bibliography{neurips_2022}
%%%%%%%%%%%%%%%%%%%%%%%%%%%%%%%%%%%%%%%%%%%%%%%%%%%%%%%%%%%%

\end{document}